\documentclass[twocolumn,10pt]{asme2ej}
\usepackage{hyperref}
\usepackage{graphicx} 
\usepackage{mathtools}
\usepackage{amsmath}
\usepackage{amsfonts}
\usepackage{subcaption}
\usepackage{xcolor}
\title{Deep Generative Models in Engineering Design: A Review}

\author{Lyle Regenwetter	
    \affiliation{
	Dept. of Mechanical Engineering\\
	Massachusetts Institute of Technology\\
	Cambridge, MA 02139\\
    Email: regenwet@mit.edu
    }
}
\author{Amin Heyrani Nobari
    \affiliation{
	Dept. of Mechanical Engineering\\
	Massachusetts Institute of Technology\\
	Cambridge, MA 02139\\
    Email: ahnobari@mit.edu
    }	
}
\author{Faez Ahmed
    \affiliation{
	Dept. of Mechanical Engineering\\
	Massachusetts Institute of Technology\\
	Cambridge, MA 02139\\
    Email: faez@mit.edu
    }	
}
\newcommand{\eg}{{\em e.g.}}
\newcommand{\etal}{{\em et~al.}}
\newcommand{\ie}{{\em i.e.}}
\newcommand{\etc}{{\em etc.}}
\newcommand{\RNum}[1]{\uppercase\expandafter{\romannumeral #1\relax}}
\begin{document}

\maketitle    

\begin{abstract}
{
{\it 
Automated design synthesis has the potential to revolutionize the modern engineering design process and improve access to highly optimized and customized products across countless industries. Successfully adapting generative Machine Learning to design engineering may enable such automated design synthesis and is a research subject of great importance. We present a review and analysis of Deep Generative Machine Learning models in engineering design. Deep Generative Models (DGMs) typically leverage deep networks to learn from an input dataset and synthesize new designs. Recently, DGMs such as feedforward Neural Networks (NNs),  Generative Adversarial Networks (GANs), Variational Autoencoders (VAEs), and certain Deep Reinforcement Learning (DRL) frameworks have shown promising results in design applications like structural optimization, materials design, and shape synthesis. The prevalence of DGMs in engineering design has skyrocketed since 2016. Anticipating continued growth, we conduct a review of recent advances to benefit researchers interested in DGMs for design. We structure our review as an exposition of the algorithms, datasets, representation methods, and applications commonly used in the current literature. In particular, we discuss key works that have introduced new techniques and methods in DGMs, successfully applied DGMs to a design-related domain, or directly supported the development of DGMs through datasets or auxiliary methods. We further identify key challenges and limitations currently seen in DGMs across design fields, such as design creativity, handling constraints and objectives, and modeling both form and functional performance simultaneously. In our discussion, we identify possible solution pathways as key areas on which to target future work. 

}
}
\end{abstract}


\section{Introduction}
The human design process is a ubiquitous element of modern society, playing a critical role in the technologies producing the food we eat, the products we use, and the spaces in which we live. Accelerating the design process through automation can reduce cost and increase industrial productivity, which would be immensely desirable for global productivity and prosperity. Integrating AI into the design process can alleviate dependence on human experts and revolutionize user customizability, providing specialized products for individual users without the prohibitive cost of manual design. Driven by the widespread potential to advance global equity and prosperity through design automation, methods such as ``generative design'' have recently emerged alongside advanced computing and automation technologies. 

``Generative design'' is the process in which algorithms directly synthesize designs either via explicit programming or implicit learning. Early generative design methods leaned heavily on explicit programming of human design expertise through manually-defined design representation methods like grammars~\cite{chakrabarti2011computer}. While practical for explicitly encoding design constraints and objectives, these rule-based frameworks ignored opportunities for implicit leaning on information and knowledge encoded in the vast expanse of existing designs.
As the availability of computational resources increased over the past decade, data-intensive methods like deep learning opened doors to successfully automate complex human tasks such as image processing and natural language processing. 

In deep learning, data is propagated through sequential layers to learn progressively higher-level meaning, an architecture generally known as an Artificial Neural Network (ANN) or just Neural Network (NN)~\cite{deng2014deep}. Most of the deep learning-based approaches pioneered during the 2010s leveraged extensive quantities of data to avoid explicit feature engineering. This trend is mirrored in engineering design with algorithms learning data distributions instead of requiring them to be predefined. Among these algorithms are Deep Generative Models (DGMs) --- deep learning models that can approximate complicated, high-dimensional probability distributions using a large dataset. In this review paper, we specifically define ``Deep Generative Models'' as algorithms that are capable of generating new samples using deep learning. Generative Adversarial Networks (GANs) and Variational Autoencoders (VAEs) are two classes of DGMs that have demonstrated compelling synthesis of images, text, and tabular data in numerous domains. Considering that images, text, and tabular data are all common representation methods for design, one might assume that DGMs should be capable of synthesizing full designs as well with relative ease. However, several unique properties of the generative design task pose particular challenges for DGMs. Many of these challenges are so fundamental that the future success of DGMs in engineering design is largely contingent on the ability to overcome them. We list four of these challenges below:
\begin{enumerate}
    \item Modeling design performance: Real-world functional performance is critical in many engineering design tasks. Developing performance-aware DGMs capable of synthesizing designs for a given set of target requirements (a process termed as inverse design) is a challenging task that is exacerbated by the computational cost of numerical simulation and the even greater difficulty of real-world evaluation. 
    \item Data sparsity: Compared to other research fields like Computer Vision, which have massive publicly available datasets, the availability of large, well-annotated, public datasets in engineering is severely lacking. Furthermore, even when data is available, the distribution of the data often does not cover the design space evenly, with much sparsity often observed in the data.
    \item The creativity gap: In conventional DGM applications, the overarching goal is to mimic the training data and emulate existing designs. In engineering design, emulation of existing products is often undesirable. Designers typically aim to introduce products with novel features to target new market segments.
    \item Usability and feasibility: For synthesized designs to be physically fabricated, they must be physically feasible. Furthermore, designs must be encoded in a data representation that contains enough parametric detail to be converted into a representation usable for fabrication. 
\end{enumerate}

Over the past few years, the design community has made substantial progress in using generative machine learning (ML) models to create new designs. DGMs have been applied to a broad range of design tasks such as structural optimization, materials design, and shape synthesis. Over time, researchers have introduced increasingly advanced methods, which have begun to address some of the above challenges. For example, many works have proposed approaches to incorporate design performance and optimization into DGM training. Other works have explored incorporating novelty and creativity into DGMs. Despite these advancements, DGMs for engineering design are still in their infancy and will require further efforts to effectively overcome these fundamental challenges. 

\begin{figure*}
    \centering
    \includegraphics[width=\textwidth]{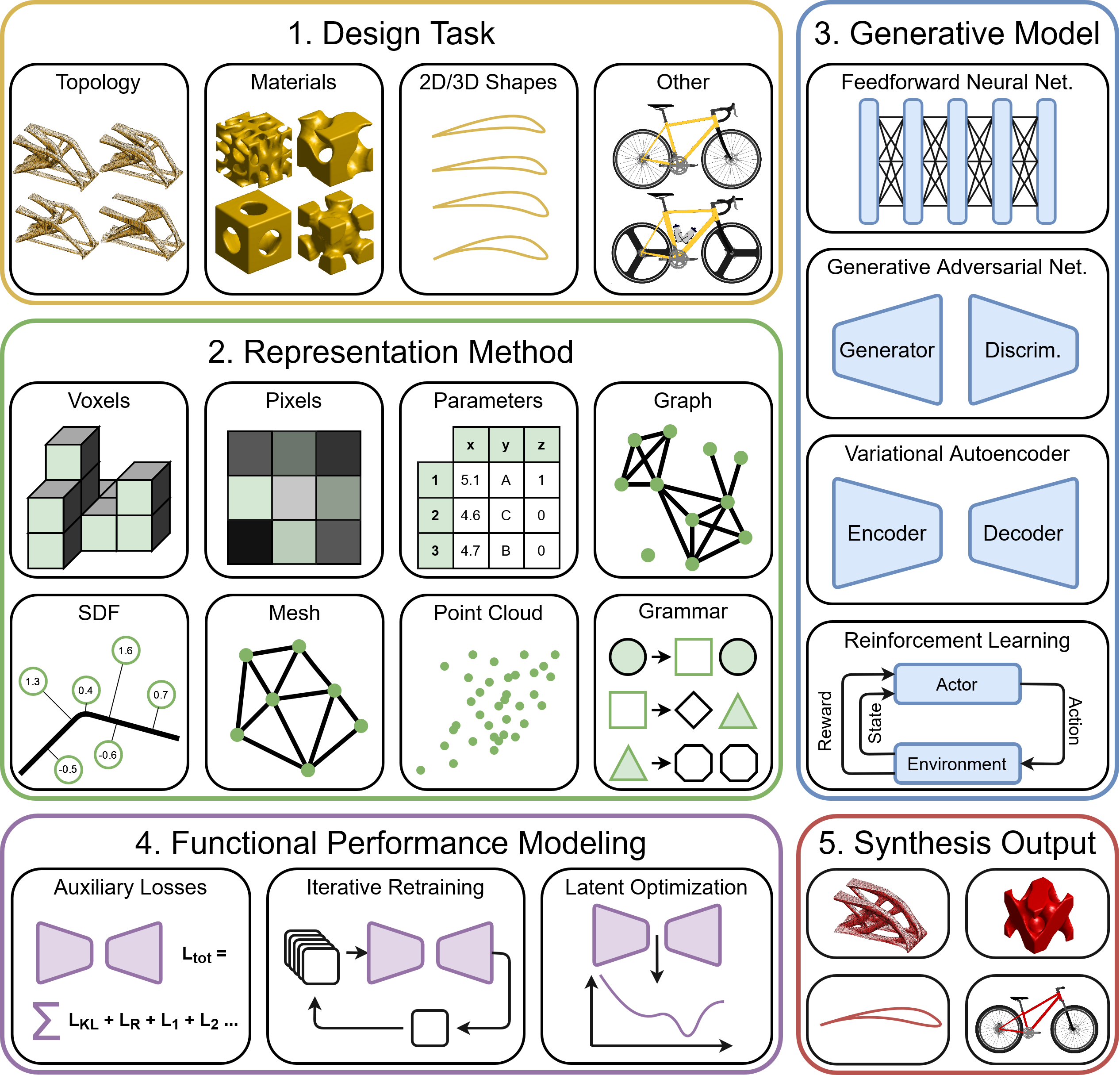}
    \caption{This figure outlines the typical components of design synthesis problems using Deep Generative Models. Some design problems are more suitable for specific design representation methods, which also influences the type of deep generative model architectures required.}
    \label{fig:DGM}
\end{figure*}

Our primary goal in this work is to help build cohesion between the countless active researchers in the design field working with DGMs and furthermore provide a starting point for researchers entering the field. 
In particular, we seek to provide researchers with a reference guide in planning projects in the data-driven generative design space. To this end, we provide an overview of common methods and tools (Sec.~\ref{methods}), a discussion of different data parameterization methods (Sec.~\ref{representation}), a review of potentially relevant research across various design domains (Sec.~\ref{sec:Domains}), an overview of relevant datasets (Sec.~\ref{sec:Datasets}), and an analysis of common challenges in the field (Sec.~\ref{discussion}). Figure~\ref{fig:DGM} provides an overview of the standard process to apply DGMs in engineering design.

\section{Overview of Deep Generative Models}
\label{methods}
Deep Generative Machine Learning approaches share the goal of high-quality synthesis but significantly vary in methodology. In practice, we identify four common approaches to generate designs: direct generation using deep neural networks~(DNN), adversarial generation with Generative Adversarial Networks~(GAN), generation from embedding vectors using Variational Autoencoders~(VAE), and sequential generation using Reinforcement learning~(RL). In the design community, we observe that GANs, VAEs, and RL are most commonly used for design synthesis. While DNNs as well as extensions like recurrent neural networks (RNNs) are occasionally used for direct design synthesis, they are more frequently used for non-generative tasks. In this section, we briefly discuss the background and methodology of GANs, VAEs, and RL.

\subsection{Generative Adversarial Networks} \label{sec:GAN}
Originally introduced in 2014, Generative Adversarial Networks (GANs)~\cite{goodfellow2014generative} found initial success with convincing image synthesis performance\cite{CycleGAN2017, choi2020stargan, karras2020analyzing,karras2019style}. We provide an introduction of GANs but refer the reader to~\cite{creswell2018generative} for a detailed overview. A generative adversarial network~\cite{goodfellow2014generative} consists of two models~\textemdash~a \textit{generator} and a \textit{discriminator}. The generator $G$ maps an arbitrary noise distribution to the data distribution, in our case the distribution of designs, and can thus generate new data; simultaneously, the discriminator $D$ learns to distinguish between real and generated data. Both models are usually built with deep neural networks. As $D$ improves, $G$ also improves as it learns to generate data that fools $D$.

\paragraph{Common challenges in training GAN models:}
GANs are often considered difficult to train, suffering from training instability stemming from several sources~\cite{salimans2016improved, arjovsky2017towards}. One common issue in GAN training occurs when the discriminator overpowers the generator, easily distinguishing generated samples, and causing the gradient to the generator to vanish, effectively halting the generator's training. This issue has been addressed by many researchers. For example, the WGAN replaces the discriminator with a critic, modifies the GAN's loss function to estimate the Wasserstein (Earth Mover's) distance between the original data and generated data distributions, and modifies the training process~\cite{arjovsky2017wasserstein, gulrajani2017improved}.

Another problem that GANs face is the issue of ``mode collapse,'' where the generator fails to encompass all modes in the data distribution or even generates only a handful of unique samples that are capable of fooling the discriminator. To overcome these issues, researchers have developed novel algorithmic techniques~\cite{VeeGAN,improvinggan,chen2021padgan} to reward diversity in samples. 

\paragraph{GAN conditioning:} \label{sec:conditioning}
In the design domain, we often have design constraints, requirements, and objectives that any generated design should satisfy. To use DGMs for such problems, these requirements should be imposed on them. For example, we may seek to train a DGM to generate bikes, but depending on our user, we may want to constrain it to generate only roadbikes or mountain bikes without retraining for each generation task. Model conditioning is one method to do this. Several proposed approaches add conditioning to the GAN using a condition vector which is intended to be interpretable. Typically GANs are discretely conditioned by feeding the condition vector into both the generator and discriminator, in a configuration known as a Conditional GAN (cGAN)~\cite{mirza2014conditional}. Instead of feeding the condition vector into the discriminator, an auxiliary network and cross entropy loss can instead be used to reconstruct the condition vector from the generated samples in a configuration known as an Information Maximizing GAN (InfoGAN)~\cite{chen2016infogan}. Conditioning is also essential in design applications where inverse design is being done on performance metrics, which often exist in continuous spaces (\eg, stiffness, lift coefficient, drag coefficient, density, \etc). Researchers have come up with continuous conditioning solutions for GANs such as the Regressional GAN~\cite{dong2019inverse}, continuous conditional GAN~(CcGAN)~\cite{ding2020ccgan} and performance conditioned diverse GAN~(PcDGAN)~\cite{pcdgan}.


\subsection{Variational Autoencoders} \label{sec:VAE}
Introduced in 2013, Variational Autoencoders found significant success in many machine learning applications. Autoencoders are unsupervised embedding algorithms consisting of an \textit{encoder} that maps an input design into a (typically) lower-dimensional latent space and a \textit{decoder} that reconstructs the design as accurately as possible from the latent space. The encoder and decoder are conventionally implemented using deep neural networks. To generate new samples, latent vectors are sampled from the latent space and fed through the decoder. Typically, the distribution of the real data mapped to the latent space of an autoencoder is sparse, meaning that sampling a realistic latent vector is difficult. This limitation is addressed with the introduction of the Variational Autoencoder (VAE), first proposed by Kingma~\etal~\cite{kingma2013auto}. The Variational Autoencoder adds in a probabilistic sampling in the latent space that regularizes the latent distribution. In practice, the VAEs's encoder outputs $n$ means and $n$ variances, from which $n$-dimensional latent vectors are sampled before decoding. To maintain a predictable latent space distribution, The VAE adds a Kullback-Liebler (KL) divergence~\cite{kullback1951information} loss between the distribution of the latent space and a standard Gaussian.
Interested readers are encouraged to refer to the literature~\cite{kingma2019introduction} for a more detailed overview. 

\paragraph{Conditional VAEs:} \label{sec:cVAE}
Just as we do for GANs, we may also seek to condition VAE training on design constraints or user preferences. The VAE has a natural advantage over the GAN in that its latent space is typically already structured. Since this structure may be fairly weak and difficult to interpret, explicitly conditioning VAEs may still be desirable. The Conditional VAE (cVAE)~\cite{sohn2015learning} extends on the conventional VAE by adding a conditioning vector as an input to both the encoder and decoder and helps achieve this goal.

\subsection{Reinforcement Learning}
Reinforcement Learning fundamentally differs from the other DGMs discussed in that it learns without a dataset in an unsupervised fashion through a large set of trial and error interactions between an \textit{actor} and an \textit{environment}~\cite{kaelbling1996RL}. This is typically done through some reward signal being sent to the actor after taking actions, based on the effects of said actions on the environment. In this scenario, the actor’s goal is to maximize the rewards it receives by making decisions (\ie, taking actions) such that the total reward is maximized. From this point of view, reinforcement learning can be thought of as an approach similar to optimization, where an objective (maximizing the reward) is being optimized.



One of the first attempts at introducing deep learning to the reinforcement learning approach was done in 2013 by Mnih~\etal, when they introduced deep learning to a reinforcement learning process known as Q-Learning~\cite{mnih2015human}. Q-Learning refers to learning the state-action value function or Q-function, which is a progressively updated estimate of the expected reward to be received from taking a particular action in a particular state. Mnih~\etal, attempted to learn the Q-function using convolutional neural networks (CNN). Many Deep RL techniques have been introduced since this first work by Mnih~\etal. Further exploration of the details of these approaches is left to the reader.



In practice, when applying RL to design applications, the design process is usually broken down into a sequential process of building a design or altering existing designs in steps~(\ie, actions taken to alter or expand the current state of a design being generated) and the reward is measured by the quality or performance of the resulting design~(\ie, the environment). While RL requires no dataset, this advantage is balanced by dependence on meaningful and reliable reward signals, which may often require a high-fidelity simulation environment. One major benefit of RL over GANs and VAEs is the fact that the reward function can be set based on any objective which does not need to be differentiable. In contrast, any objective added to the loss function of a GAN or VAE must be differentiable since GANs and VAEs are trained using the gradient-based optimization~\cite{chen2021padgan}. 

\section{Overview of Design Representation Methods} \label{representation}
In this section, we discuss common design representation methods seen in DGMs for engineering design which are visualized in Section 2 of Figure~\ref{fig:DGM}. We include a definition and discuss the pros and cons of each method.

\subsection{Images}
Design data often comes in the form of images (\eg{} microstructure scans) or can be represented in image form (\eg{} Topology Optimization). An image consists of a rectangular grid of pixels, each of which contains a color parameter. They are commonly represented by third-order tensors~($height \times width \times channels$). Common color schemes are black-and-white (boolean color channel), grayscale (integer color channel), and color (3-4 integer color channels). \textbf{Pros:} The image is an information-rich representation and can capture many details of a design. The use of convolution/convolution-transpose filters in deep learning provides a convenient tool for learning/generation of both high-level and low-level features as well as upsampling/downsampling. Many cutting-edge ML techniques are pioneered in the computer vision domain and are often directly applicable to images. \textbf{Cons:} Representing designs using pixels means that the generated design images can be infeasible for downstream tasks. Accurately fabricating designs based on images can be difficult or impossible. Even performance evaluation using conventional simulation tools like FEA or CFD can require an intermediate conversion from an image to a 3D model. The poor usability of images is exacerbated by the prevalence of artifacts in many applications (hanging pixels, disconnected geometry, \etc ). Artifacts are especially common when training on (typically) small datasets in the design domain since training is often terminated early due to over-fitting concerns. All in all, images can be considered surrogate representations of engineering designs and may lack domain knowledge and information on the physical realization of the design. Therefore, DGMs using images as representations often have a gap between the generated images and the actual design they are representing.

\subsection{Voxelizations}
Voxels are 3D grid points that are effectively the 3D equivalent of pixels. As such, voxelizations share many characteristics with images. In practice, voxels aren't conducive to `color' parameterization and are typically represented as booleans (space vs. object). This effectively makes them third-order tensors with dimension ($height \times width \times depth$). \textbf{Pros:} Voxelizations support 3D convolution which can learn high-level and low-level features in 3D. \textbf{Cons:} Compared to images and other representations, the curse of dimensionality is especially pronounced with voxels, with the number of parameters scaling with the cube of spatial resolution. Voxelizations share the same issues as images. Their usability is limited in downstream tasks and artifacts are very prevalent. Like images, voxelizations serve as surrogate representations~(often representing CAD models which originate from parametric representations or 3D shapes which originate from meshes). Like many other representations such as point clouds and Signed Distance Fields, voxelizations often require conversion before they can be used in downstream tasks. For example, they are often converted to Boundary Representation (BRep) or polygonal representations, which are often the native parameterizations of rendering and graphics software, Finite Element Analysis, and Computational Fluid Dynamics simulation. 

\subsection{Point Clouds}
Point Clouds are simple collections of points, often in 3D space, which are defined to be within some object. \textbf{Pros:} Point Clouds can represent arbitrarily complex geometry with a finite number of points, though fidelity may vary. Point Clouds are often the native output of 3D scanning software, making them relatively easy to create~\cite{daneshmand20183d}. \textbf{Cons:} Like Voxelizations and Signed Distance Functions, Point Clouds often require conversion to BRep or polygonal representations such as meshes~\cite{remondino2003point} for downstream tasks. 

\subsection{Meshes}
Meshes are a common method to represent objects in 3D space. Triangular meshes are by far the most commonly used form. Triangular meshes are the native representation used in many computer graphics algorithms and software, as well as many Finite Element tools. \textbf{Pros:} Meshes can be directly visualized and simulated in many FEA or CFD tools, enabling easy pipelines for performance evaluation using numerical methods. A mesh can be considered a specialized type of graph and can leverage graph operators like graph convolutional operators. \textbf{Cons:} In contrast to other representations like voxelizations and point clouds, meshes are more challenging to directly generate using Machine Learning methods, despite recent advances in algorithms that directly generate meshes~\cite{ranjan2018generating, cheng2019meshgan, zhang2020meshingnet}. 

\subsection{Signed Distance Functions}
The Signed Distance Function/Field (SDF) is a representation method that consists of a (typically 3D) functional map from a coordinate point to an SDF value. The magnitude of this value indicates the distance to the nearest point on the surface of the object and the sign indicates whether the point is inside or outside the object. SDFs themselves can be represented in many ways, for example, as a rasterized grid in which each `voxel' contains a continuous numerical value denoting the SDF value at that point. \textbf{Pros:} SDFs can serve as a convenient intermediate parameterization for many learning tasks. \textbf{Cons:} Like point clouds or voxels, SDFs are difficult to use in downstream tasks without first converting to BRep or polygonal representations. 

\subsection{Parameterizations}
We use the term ``parametric'' data to encompass any design representation consisting of a collection of design parameters where any spatial or temporal significance of parameters is unknown. Most parametric data can be organized in tabular form with each row being a collection of parameters representing a single design and each column describing a design parameter. Tabular design data often consists of a collection of mixed-datatype parameters where relations between these parameters may be unclear or nonexistent. Since parametric data may come in many varieties, the pros and cons discussed may not apply to every case. \textbf{Pros:} Quality parametric data is typically very information-dense (i.e. requiring fewer parameters to encode the same level of geometric detail). Whereas spatially-organized representations such as pixels or voxels encode designs with uniform information density, parametric data can contain more detail in design-critical areas without the need for upsampling the entire representation. This information density often comes with a lower dimensionality which can make optimization of parametrically represented designs significantly easier. Parametric data may also be more supportive of downstream tasks, especially if design parameters are human-interpretable. For example, a detailed enough design parameterization may allow generated designs to be directly fabricated using conventional (non-additive) manufacturing techniques. Human-interpretable parameterizations can also give human designers a tractable method to interact with generative methods to allow for human-in-loop design. Finally, design parameters can sometimes be directly linked to the latent space of a generative method, as demonstrated in numerous works~\cite{wang2020deep, chen2019synthesizing}, creating a pipeline to directly condition design generation on high-level design goals. Linking design parameters with latent variables has several potential advantages, such as enabling more effective optimization or inverse design using generative methods. \textbf{Cons:} Learning parametric data can be particularly challenging. Parametric data commonly uses mixed datatypes and inherits the training challenges of the constituent components. Multimodal distributions, skewed categories, non-Gaussian distributions, data sparsity, and poor data scaling additionally make the application of DGMs and training very difficult. Since methods that are robust to all of the mentioned challenges are difficult to come by, successfully applying existing methods to the parametric data domain can be hard. Finally, since parametric data may be nontrivial to convert to 3D models, generated parametric designs may be challenging to evaluate using numerical simulations or through qualitative visualization. 

\subsection{Grammars}
Grammars are representation methods consisting of variables, terminal symbols, nonterminal symbols, and a set of rules. Rules describe how non-terminal symbols can expand into other terminal and nonterminal symbols. The most prevalent grammars in engineering design are graph and spatial grammars~\cite{chakrabarti2011computer}, and they have been applied in a wide variety of applications, such as the design of satellites and electro-mechanical systems. Grammars can be especially useful to dictate feasible assembly hierarchies of design components as in ~\cite{stump2019spatial}. 
\textbf{Pros:} Grammars can explicitly constrain design spaces to feasible or desirable regions by nature of their construction, thereby encoding domain knowledge. \textbf{Cons:} Grammars are challenging to implicitly learn and often must be manually defined. Grammars can also restrict the exploration of the design space.
A survey on grammar-based design synthesis approaches is provided in~\cite{chakrabarti2011computer}.

\subsection{Graphs}
Graphs are a highly flexible representation method consisting of nodes and edges, which can be directed or undirected. Graph-based representations have proven successful for many different aspects of design generation and optimization and provided avenues for describing complex systems efficiently. \textbf{Pros:} Graphs are highly adaptable and are capable of representing many different kinds of complex systems and designs~\cite{10.1115/DETC2020-22355, 10.1115/1.4038303,10.1115/1.2916916,LEE1996831}. Graphs also provide representations for design processes and modeling complex interactions in systems~\cite{10.1115/DETC2014-35652,Patalano2013AGS} which may enable methods for automating the design of systems or modeling inter-part dependencies. Graph neural networks~(GNNs)~\cite{GCN,GTN,GAT,vashishth2019composition,li2017gated} provide an excellent tool for machine learning on graphs. \textbf{Cons:} Despite the developments of graph neural networks (GNNs) in the computer science community~\cite{NEURIPS2019_d0921d44,bojchevski2018netgan,you2018graphrnn,li2018learning} and their success in molecular graph generation~\cite{decao2018molgan,you2018graph}, there is less usage of graph-based DGMs in the design community, possibly due to the lack of graph-based design datasets.

\section{Literature Review Methodology}
Sec.~\ref{sec:Domains} discusses specific works that apply Deep Generative Models to engineering design or make advancements to existing generative ML methods in the context of engineering design. We consider works based on a predefined scope, with each work we discuss meeting the following selection criteria. Note that these criteria only apply to the engineering design papers presented and that the works we cite to add context to or substantiate the discussion of fundamentals, applications, and datasets need not adhere to these rules. 
\begin{enumerate}
    \item We limit our consideration specifically to papers involving Deep Generative Models, focusing on Variational Autoencoders, Generative Adversarial Networks, and Reinforcement Learning in particular. Works must utilize deep learning. Works only considering design optimization are excluded.
    \item We only consider work specifically relevant to engineering design and not other domains~(such as computer science).
    \item We consider only work published between Jan.~2014 and Sep.~2021, when we conclude our review, as many pivotal works in deep learning (CNNs, VAEs, GANs) were introduced in this period.

\end{enumerate}
To identify works, we specifically searched for ``Generative Adversarial Network,'' ``Variational Autoencoder,'' and ``Reinforcement Learning'' in Google Scholar. We initially confined our search to a set of known design venues, specifically the Journal of Mechanical Design, the Proceedings of the International Design Engineering Technical Conferences, Computer-Aided Design Journal, International Conference on Engineering Design, and Artificial Intelligence for Engineering Design, Analysis, and Manufacturing Journal. This helped us identify an initial set of seed papers. From all search results identified through these methods, 41 papers were deemed to be relevant and included. The relevance was decided by independent assessment by two raters, who are also authors of this paper. For papers where there was a disagreement, all authors discussed them and mutually decided on their classification. Next, papers cited in these seed papers were considered and added to this paper. 22 papers from other venues were deemed to be relevant and included as well. Of the 63 papers, 48 were published between Jan. 2019 and Aug. 2021, while a mere 15 were published between Jan. 2016 and Dec. 2018, indicating strong growth in the field in recent years.

\begin{table*}[]
\centering
\caption{Characterization of application studies by domain and architecture. Some works use multiple architectures or fit into multiple application domains and are listed more than once. We additionally classify works by datatype used: Image$^{i}$, Point Cloud$^{c}$, Voxelelization$^{v}$, Mesh$^{m}$, Signed Distance Field$^{s}$, Grammar$^g$, Graph$^h$, and Parametric$^{p}$ (other parameterization)}
\label{tab:Applications}
\resizebox{\textwidth}{!}{%
\begin{tabular}{|l|l|l|l|l|l|}
\hline
 & Topology Optimization  & Materials & 2D Shape Synthesis & 3D shape synthesis & Other Domains \\ \hline
Plain NN &\cite{sosnovik2019neural}$^{i}$~\cite{BEHZADI2021103014}$^{vi}$~\cite{KESHAVARZZADEH2021102947}$^{p*}$~\cite{CANG201912}$^{i}$ &\cite{malkiel2018plasmonic}$^p$~\cite{li2018transfer}$^i$~\cite{jung2021super}$^i$  &  &  &  \\ \hline
GAN &~\cite{LI2019172}$^{i}$~\cite{rawat2019application}$^{i}$~\cite{oh2018design}$^{i}$~\cite{oh2019deep}$^{i}$  &\cite{tan2020deep}\cite{yang2018microstructural}$^{i}$~\cite{ZHANG2021103041}$^{vi}$~\cite{mosser2017reconstruction}$^{v}$~\cite{lee2021virtual}$^i$~\cite{liu2019case}$^{vi}$&  &\cite{shu20203d}$^{c}$  &  \\ \hline
GAN+Conditioning &\cite{sharpe2019topology}$^{i}$\cite{nie2021topologygan}$^{i}$\cite{yu2019deep}$^{i}$~\cite{Valdez2021framework}$^{i}$ &  &\cite{yilmaz2020conditional}$^{p}$\cite{chen2018b}$^{p}$\cite{chen2019aerodynamic}$^{p}$\cite{chen2019synthesizing}$^{ph}$ &  &  \\ \hline
New GAN-based &\cite{yu2019deep}$^{i}$  & & \cite{chen2021padgan}$^{p}$ &\cite{rangegan}$^{v}$  &~\cite{wang_peng_li_chen_wu_wang_childs_guo_2019}$^{i}$\\ \hline
AE/VAE &\cite{guo2018indirect}$^{i}$  &\cite{cang2018improving}$^{i}$~\cite{li2020designing}$^i$~\cite{liu2020hybrid}$^i$~\cite{wang2020deep}$^i$~\cite{xue2020machine}$^i$ &  &\cite{brock2016context}$^{v}$  &\cite{deshpande2019computational}$^{p}$,~\cite{sharma2020path}$^{p}$,\cite{regenwetter2021biked}$^{p}$  \\ \hline
cVAE &  &\cite{tang2020generative}$^{i\,\dag}$~\cite{chen2021geometry}$^{i}$  &  &  &\cite{deshpande2019computational}$^{p}$\cite{burnap2016estimating}$^{i}$ \\ \hline
New VAE-based &  &\cite{ma2019probabilistic}$^i$  &  &\cite{zhang20193d}$^{s}$  &~\cite{10.1115/1.4048422}$^{ip}$  \\ \hline
RL-based &  &  &\cite{li2021learning}$^{p}$ \cite{dering2018physics}$^{p}$ &  &\cite{lee2019case}$^{ip}$~\cite{stump2019spatial}$^{g}$~\cite{raina2021goal}$^{ip}$~\cite{10.1115/DETC2019-97711}$^{p}$~\cite{10.1115/DETC2020-22624}$^{p}$
 \\ \hline
Other Method &\cite{greminger2020generative}$^{v}$~\cite{Fujita2021Design}$^i$ &\cite{cang2017scalable}$^{i}$~\cite{cang2017microstructure}$^{i}$~\cite{lee2021virtual}$^i$~\cite{fokina2020microstructure}$^i$~\cite{Wang2021gaussian}$^i$~\cite{10.1115/DETC2016-59404}$^i$& \cite{dering2018physics}$^{p}$  &  &
\cite{greminger2020generative}$^{v}$\cite{10.1115/DETC2018-85529}$^{p}$~\cite{stump2019spatial}$^{g}$~\cite{raina2019learning}$^{ip}$~\cite{puentes2020modeling}$^{ip}$~\cite{yukish2020using}$^g$\\ \hline 
+Style Transfer &\cite{guo2018indirect}$^{i}$  &\cite{yang2018microstructural}$^{i}$\cite{ZHANG2021103041}$^{vi}$\cite{cang2018improving}$^{i}$~\cite{li2018transfer}$^i$~\cite{fokina2020microstructure}$^i$&  &  &  \\ \hline
+Genetic Alg. & &\cite{liu2020hybrid}$^i$~\cite{Wang2021gaussian}$^i$ &  &\cite{zhang20193d}$^{s}$  &   \\ \hline

+Bayesian Opt. &  &\cite{yang2018microstructural}$^{i}$~\cite{xue2020machine}$^i$ &\cite{chen2019aerodynamic}$^{p}$  &  &  \\ 
\hline
\end{tabular}%
}
*Model-based reconstruction using parameters in a lower-dimensional space

$\dag$ Technically not images, but use an image-like parameterization structure
\end{table*}

\section{Application Domains in Engineering Design} \label{sec:Domains}
Engineering design encompasses a wide variety of applications, ranging from designing aircraft models to small-scale metamaterials.
To structure different types of applications using DGMs, we grouped them into a few categories of application areas, which are discussed in this section and are reported in Table~\ref{tab:Applications}.

\subsection{Deep Generative Models in Topology Optimization} \label{sec:TO}
\vskip -0.2 in
\begin{figure}[htb]
    \centering
    \includegraphics[width=0.99\linewidth]{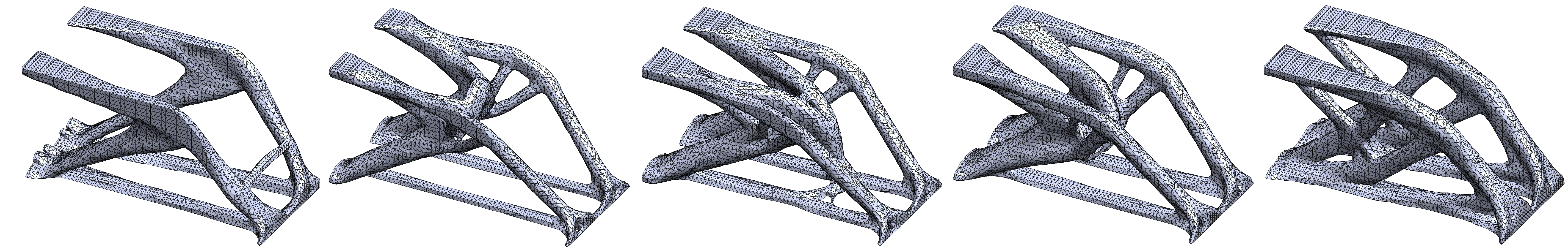}
    \caption{Sample topologies generated by 3D Topology Optimization}   
    \label{fig:topology}
\end{figure}
\vskip -0.3 in

Topology Optimization (TO) is a research field with a long history of research and methods. TO searches a design space to find an ideal spatial distribution of material to optimize some predefined objective. Common areas of application include solid mechanics~\cite{zhu2016topology, xia2017recent}, fluid dynamics~\cite{borrvall2003topology, zhou2008variational}, additive manufacturing~\cite{zegard2016bridging, langelaar2016topology}, and heat transfer~\cite{dbouk2017review, koga2013development}. While Topology Optimization is considered a method for generative design, standard TO does not leverage deep learning and is not considered a DGM. In recent years, however, several papers have proposed methods to use TO and DGMs together, in many cases to address the computational cost of TO on large amounts of data. Despite differences between architectures, we found that many DGMs for TO share certain characteristics. In particular, due to the predominant use of voxelized or pixelized representations, many hybrid TO-Generative ML models use methods originally developed for computer vision applications, such as convolutional neural networks and super-resolution~\cite{LI2019172, BEHZADI2021103014}. 

\paragraph{Supervised generation of optimized topologies:} To avoid the computational cost of Topology Optimization, DGMs have been used to predict final optimized topologies directly. This approach can also be used as an initialization technique for conventional TO, which allows conventional TO algorithms to rapidly converge, saving computational cost. We consider a baseline for DGMs in TO, in which an existing generative architecture is applied to a dataset of TO-generated designs and the network is trained to generate samples mimicking the training data. A few papers fall within this category of methods: Rawat and Chen~\cite{rawat2019application} train a WGAN on a dataset generated through TO and train an auxiliary network to additionally predict performance metrics. Sharpe and Seepersad~\cite{sharpe2019topology} expand on this baseline with a cGAN conditioned on volume fraction and load location. Guo~\etal~\cite{guo2018indirect} train a VAE with an additional style transfer~\cite{gatys2015texture, gatys2016image} loss on a TO-generated dataset for heat transfer and propose iterative strategies for targeted design optimization using the VAE's latent space. Style transfer is discussed further in Sec.~\ref{sec:Microstructure}.

\paragraph{Iterative DGM training, filtering, and human guidance in TO:} 

Oh~\etal~\cite{oh2019deep, oh2018design} propose a method that expands on the baseline generative-network-fitting by iteratively synthesizing new designs, optimizing them in TO, then dropping designs that are too similar from the full collection of designs. The authors use a modified Boundary Equilibrium GAN (BEGAN)~\cite{berthelot2017began}, which extends on the WGAN and begin training on a collection of existing TO-generated topologies. However, unlike the previously discussed generative-network-fitting approaches used in literature~\cite{rawat2019application, sharpe2019topology, guo2018indirect}, the retraining and re-optimization allows the framework to generate, optimize, and explore new areas of the design space. The proposed method is applied to the problem of wheel design, with the key motivation being to find a trade-off between the aesthetics of real designs and the structural performance of designs found through TO. The authors present strong empirical results on this wheel design problem. 

The issue of gaps in the design space of topologies is also addressed by Fujita~\etal~\cite{Fujita2021Design}, who propose an approach leveraging a Variational Deep Embedding (VaDE)~\cite{jiang2017variational}. An initial dataset is generated using TO. The proposed method sequentially identifies voids in the design space using the VaDE, decodes designs from the void, optimizes them using TO, then adds them to the training dataset. 

Although the above works have proposed methods with automated retraining steps, human input can also be injected into the training process. For example, Valdez~\etal~\cite{Valdez2021framework} propose a cGAN-based human-in-loop topology design generation framework in which the designer iteratively selects design clusters to gradually hone in on preferable designs. 

\paragraph{DGMs for TO that utilize super-resolution:}

Since one of the key limitations of Topology Optimization is computational cost, which scales with the resolution, a major focus in DGMs for TO has focused on attaining high-resolution TO-like results. A common approach uses super-resolution, which is a technique used in computer vision to convert low-resolution images to high-resolution ones. Several works expand on the baseline DGM-in-TO framework by learning from a low-resolution TO-generated dataset, then performing super-resolution on synthesized topologies, such as 
Yu~\etal~\cite{yu2019deep}, who add a cGAN for upscaling. A similar approach proposed by Li~\etal~\cite{LI2019172} uses a Super-Resolution GAN (SRGAN)~\cite{ledig2017photo} to generate high-resolution optimal topologies for heat transfer problems after generating low-dimensional topologies on a different GAN. Other researchers have proposed transfer learning for the super-resolution task instead of using GANs or VAEs for super-resolution. For example, Behzadi~\etal~\cite{BEHZADI2021103014}, train a feedforward NN-based model to predict optimized topologies directly without any discriminator and using the MSE loss. Once they have trained their model on the low-resolution data, they lock the model weights and add a few layers to the model to increase the output resolution and only train said layers to obtain high-resolution samples. This avoids the need for a large quantity of high-resolution data or the extra time required to train a high-resolution model from scratch. 
Many papers that have applied super-resolution techniques demonstrate that after initial training, their framework consistently generates near-optimal topologies many times faster than classic TO~\cite{yu2019deep, LI2019172}.

\paragraph{Improving DGM-based topology generation using physical properties:}
Several papers have succeeded in improving the baseline performance of DGMs for topology generation through the use of physical properties of the design domain. For example, Nie~\etal~\cite{nie2021topologygan} propose to adapt the cGAN architecture Pix2Pix~\cite{isola2017image} to generate synthetic topologies based on spatial fields of various physical parameters (displacement, strain energy density, and Von Mises stress) as input to the generator. Topologies generated by TO are taken as ground truth for training. The authors also propose a new generator architecture combining the Squeeze and Excitation ResNet~\cite{he2016deep, hu2018squeeze} with U-Net~\cite{long2015fully}. Cang~\etal~\cite{CANG201912} use a neural network to generate optimized topologies from loading conditions. They propose an approach to rapidly evaluate the deviation of proposed solutions from the problem's optimality conditions. They then progressively augment their dataset during training by recalculating optimal solutions (using TO) for proposed solutions that are in greatest violation of these conditions.

\paragraph{Other DGM approaches in topology generation:} 
Sosnovik and Oseledets~\cite{sosnovik2019neural} propose to use gradient information from a topology distribution to inform estimates of a final topology through a DNN. Their DNN takes both the density distribution and gradient of this density distribution from some intermediate step of a TO process to generate an estimate of the final topology without waiting for TO convergence. The authors demonstrate binary accuracy of $98\%$ when predicting the final topology after only five iterations of TO, when the full TO process would take 100 iterations to complete. 

In a different kind of approach, Keshavarzzadeh~\etal~\cite{KESHAVARZZADEH2021102947}, address the generalization problem of training for different scales and different domains. They parametrically represent shapes in both 2D and 3D using their proposed ``Disjunctive Normal Shape Model'' (DNSM)~\cite{KESHAVARZZADEH2021102947}. Using this DNSM, they create a platform for resolution-independent shape reconstruction. They then train an NN model to generate optimal topologies given boundary conditions and other problem-specific information in the DNSM space, which then can be reconstructed at any resolution using the DNSM. They demonstrate that their DNSM method overcomes the super-resolution problem across a variety of problems and domains. 

\subsection{Microstructure, Nanostructure, and Metamaterials} \label{sec:Microstructure}
\vskip -0.2 in
\begin{figure}[htb]
    \centering
    \includegraphics[width=0.99\linewidth]{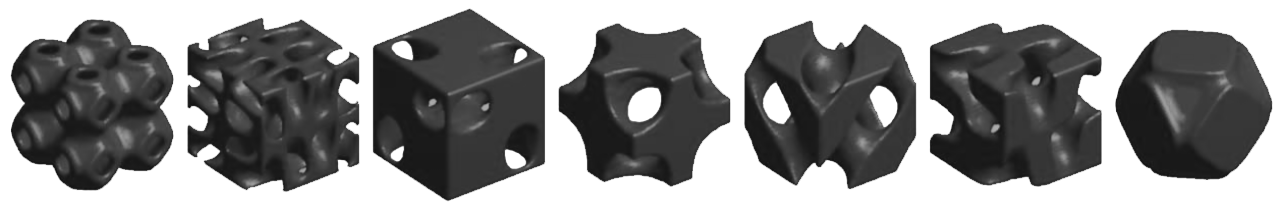}
    \caption{Deep Generative Models are often employed to generate 3D Metamaterial Unit cells}   
    \label{fig:Materials}
\end{figure}
\vskip -0.3 in

Many design applications not only require the need to design the topology or shape of the artifact, but also the material properties within it. 
Inverse design of materials is one of the key elements of Computational Materials Science. The most common approach to developing inverse materials design frameworks is the development of Process-Structure-Property (PSP) links, \ie~understanding how a particular material processing approach impacts its microstructure and how the corresponding microstructure impacts its physical properties~\cite{bostanabad2018computational}. Designing material microstructures for direct use is difficult, as there must then be some fabrication process to generate a material with the target microstructure. A common goal in computational materials science is microstructure image ``reconstruction,'' in other words, generating microstructure images that exhibit certain characteristics. Bostanabad~\etal~\cite{bostanabad2018computational} give an overview of the Microstructure Characterization and Reconstruction (MCR) field. Reconstruction can accelerate downstream tasks like augmenting the training data of networks that attempt to model PSP links. Better modeling PSP links can in turn create more accurate generative material design pipelines. Many classes of DGMs have been applied to this reconstruction task, including GANs~\cite{yang2018microstructural}, VAEs~\cite{cang2018improving} and Convolutional Deep Belief Networks (CDBNs)~\cite{lee2009convolutional}~\cite{cang2017microstructure, cang2017scalable,10.1115/DETC2016-59404}. Other studies attempt to bridge the gap between microstructure and properties in generative tasks using trained black-box surrogates, such as Tan~\etal's work~\cite{tan2020deep}. Most research focuses on 2D microstructure images, though several also consider 3D voxelizations~\cite{mosser2017reconstruction, liu2019case}.

Since reconstruction often involves mimicking existing microstructures, several papers have applied style transfer~\cite{gatys2015texture, gatys2016image} to the problem as part of a DGM architecture. Style transfer is in essence a loss between two images that attempts to capture the difference in ``style'' between the images. It is typically calculated by comparing the intermediate layers of an auxiliary style convolutional neural network. Cang~\etal~\cite{cang2018improving}, for example, propose a VAE with style transfer which is targeted at applications where only a small set of training data is available. Li~\etal~\cite{li2018transfer} also utilize style transfer in their proposed transfer learning reconstruction framework. In another approach using style transfer, this time using a vanilla GAN, Yang~\etal~\cite{yang2018microstructural} expand on the reconstruction task by attempting inverse design on the structure-property link. After training their GAN with style transfer loss and an additional loss to penalize mode collapse, they treat the noise vectors as input variables and optimize them with Bayesian optimization, using the generator to translate between design vectors and microstructure images. The authors choose to optimize microstructures for energy absorption which can be evaluated from images using coupled-wave analysis~\cite{yu2017characterization}, providing them a convenient structure-property link that would be more challenging if optimizing for other objectives. 

More recently, Fokina~\etal~\cite{fokina2020microstructure} adapt StyleGAN~\cite{karras2019style} to the microstructure image synthesis domain. Other studies impose the ``style'' of generated microstructure images through enforcement of physical properties. For example, Chen~\etal~\cite{chen2021geometry} propose an approach to generate Random Heterogeneous Material (RHM) microstructure images using a cGAN conditioned on target images. Their cGAN uses an augmented loss based on the matching of perimeter, volume, and Euler characteristics. DGM studies in microstructure design have also used other advanced methods introduced in computer vision such as super-resolution\cite{jung2021super} and image translation~\cite{lee2021virtual, CycleGAN2017 ,lee2021virtual, isola2017image}. 

Although 2D images are a widely used means to represent microstructures, Zhang~\etal~\cite{ZHANG2021103041} expand the microstructure reconstruction task to the 3D domain. Their ScaffoldGAN method generates scaffold materials to mimic real-world examples of human bone scaffolds as well as foam metal scaffolds in both 3D and 2D data. They employ the conventional GAN approach with style loss for better visual similarity between generated scaffolds and real ones. However, the authors additionally introduce a novel ``structural loss'' term to specifically address spatial coherence, a limitation of GANs in emulating scaffolds. This additional loss helps them achieve better results that mimic the features of the data more realistically. Other studies also consider DGMs on 3D microstructures, such as Mosser~\etal~\cite{mosser2017reconstruction} and Liu~\etal~\cite{liu2019case}.

\paragraph{Photonics and phononics:}
Within microstructure design, the design of materials with photonic or phononic properties is an area of interest in numerous industries including sensing, communications, and display technology.
The work by Yang~\etal~\cite{yang2018microstructural} is an example of a subfield of generative microstructure design targeting the development of microstructures with particular photonic or phononic properties. Molesky~\etal~\cite{molesky2018inverse} also provide a review of inverse design in nanophotonics. 

In the photonics and phononics subfield, performance evaluations such as the one used in~\cite{yang2018microstructural} are frequently used to incorporate performance into DGMs. For example, the technique of optimization using learned mappings from the latent space of a trained autoencoder to the property space has been employed in several papers. Li~\etal~\cite{li2020designing}, Liu~\etal~\cite{liu2020hybrid}, and Wang~\etal~\cite{Wang2021gaussian} train an autoencoder, VAE, and Gaussian Mixture VAE~\cite{dilokthanakul2016deep} respectively on images of microstructures. They map the latent variables to the property space using a DNN, CNN, and Gaussian Process Regressor, respectively. Li~\etal~\cite{li2020designing} directly optimize the properties using the DNN, while Liu~\etal~\cite{liu2020hybrid}, and Wang~\etal~\cite{Wang2021gaussian} optimize using a Genetic Algorithm. 

Though optimizing latent variables is a common approach, several other DGMs have been proposed for photonics/phononics design that optimize performance in other ways. For instance, Malkiel~\etal~\cite{malkiel2018plasmonic} uniquely use a parametric representation and a bidirectional DNN to simultaneously learn a bidirectional mapping between nanostructures and the property space. 
Ma~\etal~\cite{ma2019probabilistic} propose a novel VAE-based framework consisting of feature extraction, prediction, recognition, and generation networks. The proposed method is capable of both forward prediction of properties based on metamaterial structure as well as the inverse (generative) prediction task based on properties. Furthermore, the framework supports self-supervised learning where the model trains on unlabeled metamaterial pattern images without corresponding property labels. The authors also demonstrate transfer learning to other microstructure shape classes as well as the inverse design of multiple microstructures forming a meta-mirror with desired properties. DGMs have also been applied to nano-scale photonic devices, such as an optic broadband power splitter as in Tang~\etal~\cite{tang2020generative}. 

\paragraph{Unit-cell-based metamaterials:}
Research in generative design for microstructures has also focused on the development of unit cell structures for metamaterials. For instance, 
Wang~\etal~\cite{wang2020deep} fit a VAE to metamaterial unit cells and demonstrate that latent space parameters can reflect the physical properties of the unit cells. They then demonstrate several downstream tasks using their VAE, such as diverse subset selection and generation, targeted generation to match desired stiffness matrix values, and metamaterial family design. The authors also demonstrate several 2D macro-optimization runs to design arrays of unit cells to match macro-level deflection targets in an approach similar to classic TO. Xue~\etal~\cite{xue2020machine} train a VAE to generate unit cells, then perform Bayesian Optimization in the latent space of the VAE to attain desired macroscopic elastic properties. 

\subsection{2D Shape Synthesis} \label{sec:Geometry}
\vskip -0.2 in
\begin{figure}[htb]
    \centering
    \vskip -0.3 in
    \includegraphics[width=0.99\linewidth]{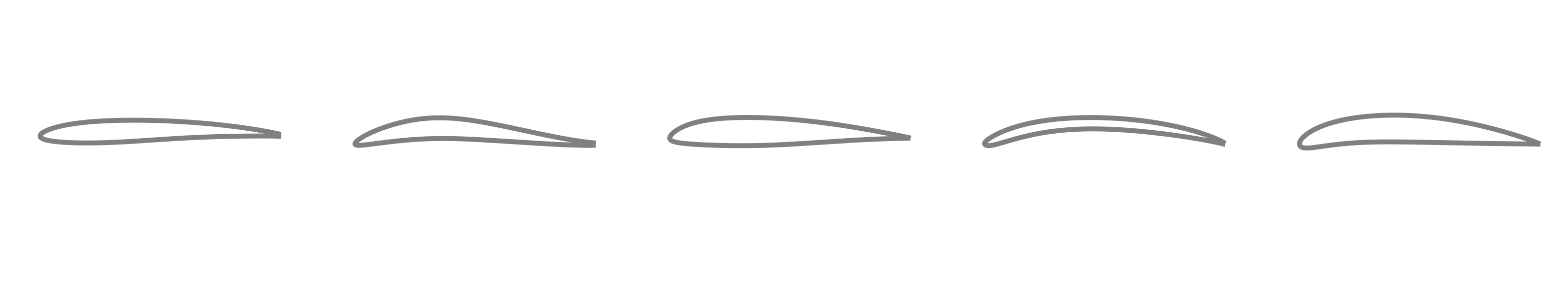}
    \vskip -0.3 in
    \caption{Airfoil synthesis using DGMs usually entails employing 2D shape generation approaches.}   
    \label{fig:Airfoils}
\end{figure}
\vskip -0.3 in
Designing features with critical geometric considerations is a task that appears in several engineering fields such as aerospace and automobile design. DGMs have been widely applied to these problems. Within aerospace, the design of an airfoil, which is the cross-sectional shape of a wing, is of particular interest to many researchers. Airfoils have a wide variety of uses in engineering domains such as propeller, rotor, and turbine blade design. Since airfoil performance parameters are of key interest, most of the research focuses on performance-aware conditional generation. For example, Yilmaz and German~\cite{yilmaz2020conditional} apply a cGAN to the airfoil design problem, conditioning their network on various stall parameters. While the overwhelming majority of research in this domain uses DGMs trained to learn shape parameterization based on spline interpolation of Cartesian points, equation-based parameterization is also sometimes used. For example, Li~\etal~\cite{li2021learning} propose a performance-aware RL framework where the RL agent learns the optimal equation coefficients using Proximal Policy Optimization (PPO)~\cite{schulman2017proximal}. 

Other works generalize their applications to general 2D shape synthesis. For example, Chen and Fuge~\cite{chen2018b} propose B{\'e}zier GAN, a framework that learns shape representations through B{\'e}zier curves, featuring InfoGAN style conditioning~\cite{chen2016infogan}. 
Chen~\etal~\cite{chen2019aerodynamic} expand on this work with Bayesian Optimization to maximize lift/drag ratio. Classic test data for 2D shape synthesis methods are the UIUC airfoil dataset as well as artificially-designed shapes like superformulas~\cite{gielis2003generic}. While the emphasis of certain papers is on shapes or airfoils themselves, several papers primarily propose methodology advancements that they choose to demonstrate on the 2D shape synthesis domain. For example, Chen and Fuge~\cite{chen2019synthesizing} address the challenging problem of multi-component design generation using a GAN to synthesize parts using inter-part dependencies. The authors assume inter-part dependencies in a design are known and propose modeling them using directed acyclic graphs. They propose a hierarchical adaptation of the InfoGAN using a single discriminator for the entire design and one generator/auxiliary network pair for each part in the design, which they term the hierarchical GAN (HGAN). The method is tested on a medley of synthetic datasets generated using B{\'e}zierGAN~\cite{chen2018b} and demonstrated to form meaningful and interpretable latent spaces. Although the paper assumes that part dependency graphs for the object class are well defined, this paper takes a significant step towards multi-component design synthesis and interpretable GANs.

Several other papers use the 2D shape synthesis domain to address the challenges of performance evaluation in DGMs. Dering~\etal~\cite{dering2018physics}, for example, apply an iterative retraining approach to boat sketches using an adaptation of the Long Short-Term Memory (LSTM)-based Sketch-RNN~\cite{ha2018neural} as the generator. To evaluate candidate designs, the performance is scored using a simulated environment in a game engine, within which the ``behavior'' (motion) of the design is learned.

A recent work by Chen and Ahmed~\cite{chen2021padgan} addresses both performance evaluation and design novelty through their proposed Performance Augmented Diverse GAN (PaDGAN) framework. Conventional GANs are trained to mimic the design space they are trained in and as a consequence, are penalized for generating novel designs. PaDGAN expands on the conventional GAN architecture by modeling design performance and design diversity using a Determinantal Point Process (DPP) kernel~\cite{kulesza2012determinantal}. By promoting diversity, PaDGAN directly addresses mode collapse in GANs too (see Sec.~\ref{sec:GAN}). The authors demonstrated that PaDGAN is capable of generating high-quality and previously unseen novel designs for the UIUC airfoil database, using the lift to drag ratio as the quality metric. \cite{chen2021mopadgan} extends this work to multi-objective generation. 

\subsection{3D Shape Synthesis} \label{sec:3DObjects}
\vskip -0.2 in
\begin{figure}[htb]
    \centering
    \includegraphics[width=0.99\linewidth]{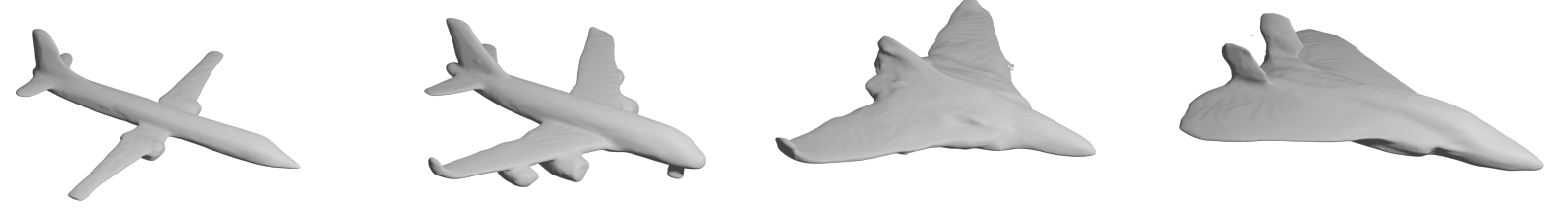}
    \caption{DGM models, such as Range-GAN~\cite{rangegan}, can be used to generate 3D models of aircrafts with constraints.}   
    \label{fig:Unitcells}
\end{figure}
\vskip -0.3 in

3D object generation using deep learning is an active field in computer science, with significant research efforts being dedicated to generating realistic-looking shapes and objects. 3D shapes and objects are typically represented by voxels, point clouds, or meshes. Advancements in 3D shape synthesis in the computer graphics community have leaned heavily on GANs or Autoencoders, as well as other machine learning advancements like Recurrent Neural Networks (RNNs), Transformers, and Graph Neural Networks (GNNs). This research is tangentially relevant to engineering design but is typically focused more on visual appearance and aesthetics rather than design considerations like functional performance derived from simulations or manufacturability. For this reason, we do not discuss these methods in detail and opt to list a few of the many noteworthy papers in this field from 2016 to 2021 for the reader to further explore at their discretion:~\cite{li2017grass, zou20173d, groueix2018papier, gao2019sdm, mo2019structurenet, park2019deepsdf}. It is important to note that many of these proposed DGM architectures could be adapted to the design domain too.
Despite the prominence of 3D object generation papers in the computer graphics community, a few have arisen within the engineering design community as well. In an early work, for example, Brock~\etal~\cite{brock2016context} implement a 3D cVAE to reconstruct and interpolate between voxelized models from the ModelNet-10 Dataset. 

The majority of 3D shape synthesis work in engineering design, however, has implemented some kind of design performance consideration. Zhang~\etal~\cite{zhang20193d} propose the use of a Genetic Algorithm (GA) to optimize latent space design embeddings of a trained VAE variant called the Variational Shape Learner (VSL)~\cite{liu2018learning}, which they demonstrate on the optimization of 3D aircraft models. Shu~\etal~\cite{shu20203d} take an iterative retraining approach by retraining a GAN on high-performance models evaluated using Computational Fluid Dynamics (CFD) evaluation. The authors apply the proposed method to point cloud aircraft models from ShapeNet and use minimization of aerodynamic drag as the performance objective of choice. The authors use a standard GAN loss and use a discriminator architecture from~\cite{qi2017pointnet}. Nobari~\etal~\cite{rangegan} introduce another self-augmentation approach to train on skewed or sparse datasets. They couple it with a ``range loss'' to encourage designs to comply with design constraints related to parameter bounds and apply their approach to generate 3D aircraft models. 

\subsection{Other Applications} \label{sec:OtherApps}
In this section, we discuss several application domains for DGMs with only a few relevant papers identified in our search. 

\paragraph{Manufacturability:} \label{sec:manufactuability}
Synthesizing manufacturable designs is an important, albeit less-explored area in DGMs.
Greminger~\cite{greminger2020generative} proposes approaching the problem of manufacturability using an MSG-GAN architecture~\cite{karnewar2020msg} adapted to the 3D domain. The author synthetically generates topologies that are manufacturable by a 3-axis mill, trains the GAN to generate similar topologies, then optimizes generated topologies using TO. A key challenge in GAN-based synthesis of manufacturable designs is a lack of annotated datasets with design and manufacturing process details.

\paragraph{Fluid flow shapes:}
DGMs have also been used to generate flow shapes. For example, Lee~\etal~\cite{lee2019case} apply a Double Deep Q-Network (DoubleDQN)~\cite{van2016deep} RL framework to learn the geometry of a microfluidic channel to generate target flow shapes. The authors also propose methods in which a human designer can participate in the design process alongside the RL agent to impart design knowledge or constraints. 
Other approaches that integrate humans into the design generation process are discussed next.

\paragraph{Integrating humans into the generation process:}
One element of design that is often less emphasized in DGMs for design is human interpretation and design input during the generative process. Though datasets are often created using some human input, humans are not directly involved during the training process of most DGMs. Burnap~\etal~\cite{burnap2016estimating} note that numerical performance measures often do not correspond to a human's perception of design quality, and study this phenomenon on automobile images generated through a conditional VAE. Supporting the involvement of humans during the training process is a potential approach to find the middle ground between the expertise and intuition of humans and the detailed information of datasets, which can enable different tasks. For example, Wang~\etal~\cite{wang_peng_li_chen_wu_wang_childs_guo_2019} look at the problem of generating samples that are visually pleasing for any given human subject. To do this, they train a modified version of the Auxiliary Classifier GAN~(ACGAN)~\cite{odena2017conditional} which classifies the type of image being generated at the same time as learning to generate images. The authors, however, do not opt for a cGAN architecture. Instead, they condition the GAN using Electroencephalography (EEG) signals from human subjects. They train an encoder that takes EEG signals and transforms them into a set of input features for the GAN. By exposing human subjects to the images as they are being generated, the encoder introduces the human response to the design into the generation process. This effectively allows for a human subject to extract more visually pleasing samples from the GAN model. Several other studies such as~\cite{Valdez2021framework, deshpande2019computational, lee2019case} propose frameworks that incorporate humans into the design process. 

\paragraph{Kinematic synthesis:}
In machine design, generating a mechanism to generate specific motion curves or transmit power is a common design task. Kinematic synthesis involves selecting the size, type, and configuration of mechanism components to achieve such a goal. Several works have applied DGMs to various kinematic synthesis problems. 
Deshpande and Purwar~\cite{deshpande2019computational}, for example, propose a VAE-based method for planar linkage synthesis. They deterministically parameterize both coupler paths and linkages, then generate sample coupler paths from a variety of linkages (four-bar, slider-crank, Stephenson-type six-bar). In the first of their two studies, the authors train a VAE to reconstruct trajectories. They then propose a method in which humans can interact with the trained VAE to customize linkage paths to their design goals by visualizing the effect of various latent space perturbations and selecting one that matches their design vision. The authors then expand on their first study with a cVAE architecture, feeding in the linkage as the data samples and the coupler curve as the condition vector. In a similar work, Sharma and Purwar~\cite{sharma2020path} explore spatial linkage synthesis of 5-SS mechanisms using VAEs. In other papers, the same authors take a look at paths themselves as guiding mediums for linkage mechanism design by first training models to encode paths, then using said models to search within a dataset of mechanisms to obtain solutions for the specific paths~\cite{10.1115/1.4048422}.

Other studies have approached the kinematic synthesis task through Reinforcement Learning. Vermeer~\etal~\cite{10.1115/DETC2018-85529} propose using Deep Q-Learning to synthesize planar linkage mechanisms to obtain mechanisms that are capable of producing straight lines, showing that RL can be effective in generating linkage mechanisms using machine learning.

\paragraph{Truss design:}
Raina~\etal~\cite{raina2019learning} propose an RL-based approach to truss design. In their approach, they learn from a dataset of sequential human design decisions~\cite{mccomb2018data}. They first train an Autoencoder to map trusses to and from a design embedding. Sequential design embeddings are then used to predict a heatmap of possible next design states using a supervised transition network. Interestingly, the authors choose to convert to and from images from their parametric representation when training the autoencoder and using decoded results. Finally, a rule-based agent selects from possible next design steps. Even without knowledge of design metrics and performance, the RL agents were found to generate competitive designs compared to humans. Puentes~\etal~\cite{puentes2020modeling} expand on this work by learning action-sequence heuristics instead of individual design actions. Raina~\etal~\cite{raina2021goal} further expand on this work with a goal-based reinforcement learning agent.

\paragraph{Hierarchical product design synthesis:}

Some works have applied DGMs to multi-component products, typically with a well-known hierarchical structure. Stump~\etal~\cite{stump2019spatial}, for example, propose a unique approach that simultaneously combines RL, Recurrent Neural Networks (RNNs), grammars, and physics-based simulation. In their approach, the RNN learns to generate modular sailing crafts by sampling discrete selections from a predefined shape grammar. The training of the RNN is detailed in~\cite{yukish2020using}. The control policy of the craft is then optimized using RL in a physics-based simulated sailing environment. Regenwetter~\etal~\cite{regenwetter2021biked, regenwetter2022biked} explore the full synthesis of bicycles, a diverse class of products typically consisting of numerous hierarchical components. The authors demonstrate full generative synthesis of bikes using VAEs for both image and parametric representations. The authors note that their detailed design parameterization allows AI-generated designs to be physically fabricated. 

\paragraph{Procedural content generation:}
Lopez~\etal, explore procedural content generation in a specific context for virtual reality. In their work, they introduce a reinforcement learning model that can learn to generate 3D virtual reality~(VR) content which users can explore in a VR environment~\cite{10.1115/DETC2019-97711}. In their approach, they specifically work towards generating manufacturing environments that are physically valid and feasible. In an extension of this work, Cunningham~\etal, extend the method to generate content for multiple contexts instead of just one at a time~\cite{10.1115/DETC2020-22624}, which enables the integration of user-specific parameters.


\section{Datasets} \label{sec:Datasets}
In this section, we present commonly-used datasets that have been or have the potential to be used to train DGMs for data-driven design tasks. We provide a more detailed list online\footnote{\url{https://decode.mit.edu/datasets/}}. While the datasets listed are not comprehensive, we aim to note key types of datasets that have been commonly used in engineering design applications. We hope that researchers can create larger well-annotated datasets and make them public for the community to use.
\subsection{Topology Optimization Datasets}
Most papers on DGMs for Topology Optimization generate their own datasets, which are often not publicly available. The most common method for dataset generation is Solid Isotropic Material with Penalisation (SIMP), which has several publicly available software implementations. A few TO datasets, however, are open source. Sosnovik and Oseledets~\cite{sosnovik2019neural} generate a dataset of 10,000 artificially generated topologies providing both final and intermediate topologies from the optimization process\footnote{\url{https://github.com/ISosnovik/top}}. Each included topology contains one 40x40 image of the topology generated at every stage of 100 steps of Topology Optimization. Hence, a total of one million images are provided. Nie~\etal~\cite{nie2021topologygan} provide the dataset used to train their TopologyGAN framework\footnote{\url{https://github.com/zhenguonie/2020_TopologyGAN}}. Unlike the aforementioned, this dataset consists only of final topologies but contains 49078 generated topologies at a resolution of 64x128, generated from 42 unique boundary conditions. Both datasets discussed use the ToPy~\cite{hunter2017topy} implementation of SIMP. 

\subsection{Microstructure Datasets}
Well-established technologies such as optical microscopy (OM) and scanning electron microscopy (SEM) are often used to visualize material microstructures. As such, numerous datasets of microstructure scan images are publicly available, such as~\cite{iren2021aachen},~\cite{larmuseau2020compact}, and ~\cite{decost2017uhcsdb}. Numerous datasets are also available for the design of composite materials of various types, such as the NanoMine nanopolymer composite database\footnote{\url{https://github.com/tetherless-world/nanomine-ontology}}, which contains over 20,000 datapoints~\cite{zhao2018nanomine}. Compiled lists of materials science datasets\footnote{\url{https://github.com/sedaoturak/data-resources-for-materials-science}} and synthetic microstructure datasets are also available. For example,
Yang~\etal~\cite{yang2018microstructural} provide a trained GAN model which generates synthetic microstructure images\footnote{\url{https://github.com/zyz293/GAN_Materials_Design}}.

\subsection{Design Geometry Datasets} The UIUC airfoil database\footnote{\url{https://m-selig.ae.illinois.edu/ads/coord_database.html}} has been used as a case study in several generative design research frameworks~\cite{chen2019aerodynamic, chen2021padgan, chen2018b}. The database details nearly 1,600 real-world airfoil designs using coordinates of points on the surface. Since the original data provides inconsistent numbers of coordinates along the top and bottom surfaces, Chen~\etal~\cite{chen2019aerodynamic} propose a method to standardize the data with B-spline interpolation over the airfoil which is also used in other works~\cite{chen2019synthesizing, chen2021padgan}. 

\subsection{3D Object Datasets}
ShapeNet\footnote{\url{https://shapenet.org/}}~\cite{chang2015shapenet} is one of the most commonly-used 3D model datasets, consisting of over 51,300 3D models of 55 object categories. PartNet~\cite{mo2019partnet} expands on ShapeNet with fine-grained hierarchical semantic annotations for component parts of ShapeNet objects. Princeton ModelNet\footnote{\url{https://modelnet.cs.princeton.edu/}}~\cite{wu20153d} is another commonly-used 3D model dataset consisting of 127,915 voxel-based 3D models of 662 object categories. Numerous works discussed in this review use ShapeNet or ModelNet models~\cite{shu20203d, rangegan, zhang20193d, brock2016context}.

Sangpil~\etal~\cite{sangpil2020large} introduce a dataset of 58,696 models of mechanical components from 68 classes called the Mechanical Components Benchmark (MCB)\footnote{\url{https://bit.ly/3ne4gwv}}. The MCB contains a hierarchical label tree grouping components into subclasses of different levels, such as Components $\rightarrow$ Fasteners $\rightarrow$ Nuts $\rightarrow$ Wingnuts, for example. Models are represented as point clouds, voxels, and 2D views. 

\subsection{CAD and CAD-based Datasets}
Willis~\etal~\cite{willis2021fusion} introduce two datasets of Autodesk Fusion models\footnote{\url{https://github.com/AutodeskAILab/Fusion360GalleryDataset}}. One dataset, intended for reconstruction tasks, contains 8,625 models and the other, intended for segmentation, contains 35,680 models. The reconstruction dataset is particularly interesting for generative tasks, as it contains information governing the sequential CAD operation steps taken to generate a part. 

Regenwetter~\etal~\cite{regenwetter2021biked, regenwetter2022biked} introduce a dataset called BIKED\footnote{\url{https://decode.mit.edu/projects/biked/}} consisting of mixed data extracted from 4,512 bicycle CAD models. The dataset includes bicycle assembly images, segmented subcomponent images, as well as ``parametric'' data. The ``parametric'' data consists of 2,395 mixed-type design parameters describing both high-level and low-level characteristics. BIKED provides an advantage over conventional 3D object datasets for generative tasks in that synthesized designs contain the necessary parametric information to physically fabricate designs. 
BIKED's parametric data is used in~\cite{regenwetter2021biked} for bicycle synthesis and its image data in~\cite{nobari2021creativegan} for generating novel designs. The FRAMED dataset~\cite{regenwetter2022framed} expands on BIKED with structural performance data, such as weight, safety factors, and deflections under various loads for all 4512 models as well as artificially generated bicycle frames\footnote{\url{https://decode.mit.edu/projects/framed/}}.

\subsection{Metamaterials Datasets}
Wang~\etal~\cite{wang2020deep} introduce a dataset of 248,396 2D unit cells represented by 50x50 pixelated matrices. The unit cells have associated stiffness tensor components provided and can also be used for TO research.  Chan~\etal~\cite{chan2021metaset} introduce a dataset of 3,000 3D isosurface unit cells sampled from 30 level-set functions, along with corresponding 3D elastic tensor components. Wang~\etal~\cite{wang2021data} introduce a dataset of 795 unit cells generated from 10 lattice models. Associated stiffness tensors are provided. The three datasets can be found here\footnote{\url{https://ideal.mech.northwestern.edu/research/software/}}.

\subsection{Sketch Datasets}
QuickDraw~\cite{jongejan2016quick} is a sketch dataset of 50 million doodles from 345 categories\footnote{\url{https://github.com/googlecreativelab/quickdraw-dataset}}. The doodles were collected by Google from user-drawn sketches in an interactive sketching game. QuickDraw data is used in several works discussed~\cite{lopez2018human, dering2018physics}. Toh \& Miller~\cite{toh2013exploring} introduce a dataset of 934 innovative milk frother design sketches with associated text descriptions that can potentially be used to train DGMs, perhaps factoring in Natural Language Processing (NLP)\footnote{\url{https://sites.psu.edu/creativitymetrics/2018/07/18/milkfrother/}}. 

\subsection{Sequential Human Design Datasets}
McComb~\etal~\cite{mccomb2018data} provide a tabular truss design dataset taken from a truss design activity executed by sixteen human teams\footnote{\url{https://www.sciencedirect.com/science/article/pii/S2352340918302014?via\%3Dihub}}. Sequential design operations such as joint and member placement are recorded using geometric parameters such as joint coordinates, member size, \etc. Performance metrics such as safety factors and weights are also included. While this dataset can be used as a truss design dataset, it is primarily intended as a resource to study or mimic the human design process. We note that the Autodesk Fusion reconstruction dataset mentioned previously~\cite{willis2021fusion} can also be used for learning sequential design tasks.

\section{Discussion, Challenges and Future Work} \label{discussion}

When applying DGMs to engineering design, the `standard' objective of mimicking the training data is often insufficient, or even counterproductive. Instead, real designs are governed by specific objectives and constraints, often including novelty or creativity. Thus, different objectives, such as real-world performance metrics, novelty, and adherence to constraints may make for better training objectives. We discuss the challenges with performance evaluation, constraint violation, and incorporation of novelty in the following sections, as well as pathways to potential solutions. We further discuss some general challenges for DGMs in engineering design, such as limited availability of data, lack of benchmark problems and metrics, and delayed adoption of cutting edge methods from different research communities. Our discussion is supported by observations from our literature review. 

\subsection{Design Performance Evaluation} \label{performance}


Incorporating performance evaluation into DGMs is one of the key stepping stones toward practical applications of ML in design. Three major challenges in performance evaluation are fidelity, cost, and differentiability. 
\begin{enumerate}
    \item Evaluation methods lacking fidelity may result in DGM-generated designs that do not meet specifications. 
    \item Computational cost precludes compatibility with methods that heavily sample performance values. 
    \item Lack of differentiability makes implementation into the training process difficult in any machine learning models which rely on gradient-based optimization. 
\end{enumerate}
Physical evaluation, which means building a product and testing its performance in the real-world, typically has the highest fidelity but is rarely adopted in DGMs due to its prohibitive cost. Qualitative human evaluation of designs has also been investigated in several papers~\cite{deshpande2019computational, Valdez2021framework, lee2019case}], although the evaluation is typically not focused on performance. Medium-fidelity evaluations, such as numerical simulations often deliver satisfactory fidelity, but can be costly and are rarely differentiable. Methods that do incorporate medium-fidelity performance evaluation like~\cite{shu20203d, oh2019deep} do not rely on performance score gradients, and instead alternate (re)training and performance evaluation, typically doing this retraining a handful of times. 

Low-cost evaluation methods like surrogate models can be worked into the training objective and are by far the most common performance evaluation method seen in DGMs~\cite{chen2021padgan,pcdgan,rangegan,dong2019inverse, sharpe2019topology, guo2018indirect, rawat2019application}. Unfortunately, surrogate models can be brittle and generalize poorly to designs that differ from the data the surrogate was trained on. This is a particular concern when training data for surrogates is unevenly distributed across the performance space~\cite{rangegan} and when novelty and performance are incorporated into training objectives for DGMs. Many different approaches have attempted to improve the performance of low-fidelity surrogate models. For example, self-supervised data augmentation~\cite{rangegan} uses the DGM itself to generate samples in sparse regions~(through optimization~\cite{chen2021mopadgan} or conditioning~\cite{rangegan}) which then can be used to train the low-fidelity models to perform more accurately and in turn improve methods that rely on them for design generation. Multi-fidelity modeling is another promising approach, which involves generating surrogate models that augment a few costly high-fidelity samples with low-fidelity samples to attain higher fidelity surrogates with minimal expense. 

Advancements have also been made in using machine learning to improve and accelerate medium-fidelity physics simulations such as Finite Element Analysis~\cite{FEA1,FEA2,FEA3,FEA8,CFD10} and Computational Fluid Dynamics~\cite{CFD1,CFD3,CFD5,CFD10}. These works, although not yet robust enough to be easily applied as an alternative to high-fidelity simulations, provide a proof of concept for machine learning-based acceleration and even replacement of higher-fidelity simulations. 

Numerous studies have also proposed crowdsourcing methods to evaluate synthesized data from generative machine learning methods~\cite{burnap2016estimating, dering2017generative, lopez2018human} instead of physical or physics-based evaluation. Though the fidelity of crowdsourced evaluation is highly task- and crowd-expertise-dependent, the high evaluation cost and lack of differentiability are fairly universal. 

In summary, current DGMs are largely constrained in performance evaluation by an inherent fidelity vs. cost tradeoff, however, several directions show promise in enabling faster and more accurate evaluation methods that may escape this limitation. 


\subsection{Feasibility, Constraints, and Manufacturability}
One key component of design performance is obeying explicit design constraints, as well as implicit constraints such as physical feasibility and manufacturability. DGMs are difficult to rely upon for explicit design constraints since they are generally probabilistic and may generate completely invalid designs. This issue is a concern that many researchers have pointed to~\cite{guest_editorial, chen2019synthesizing, cang2017microstructure}. One potential solution is to develop inexpensive and reliable validation methods, however, this is a challenging task that may require significant human input.

Another underlying issue lies in design representation, which is discussed in detail in Sec.~\ref{representation}. Representations such as images often have no clear translation to representations with practical uses. Other representations, such as 3D models of various types can only be realistically fabricated using additive manufacturing, which precludes many design domains. For ML-generated designs to be physically fabricated and used in practical applications, the feasibility across this `domain gap' must be overcome~\cite{cang2017microstructure,regenwetter2021biked}. Though a few works have investigated manufacturability~\cite{greminger2020generative}, we observed that modeling it is not considered in most of the papers we reviewed. Other works have attempted to apply DGMs to parameterizations that encode similar parametric design data to what would be used in manufacturing drawings. However, the same works identify that parametric data of this sort is challenging to learn and generate~\cite{regenwetter2021biked}. 
\subsection{Creativity and Novelty}
Whereas creativity and novelty are essential aspects of the classic design process, DGMs rarely explicitly consider either. Most DGMs learn to mimic the data covering the existing and already explored portions of the design space. While this emulative behavior is helpful for maintaining realism and ensuring sample quality, it incentivizes DGMs against generating creative or novel designs~\cite{elgammal2017can}. For DGMs to progress towards more human-like design, advancements must be made in modeling creativity as well as in developing architectures that promote creativity.


Several recent works~\cite{chen2021mopadgan, chen2021padgan, nobari2021creativegan} have proposed methods to encourage creativity and novelty in DGMs. In their framework,~CreativeGAN, Nobari~\etal, focus on identifying novelty and guiding DGMs towards such behavior by directly introducing novel features into typical designs, thereby expanding the design space and novelty of the DGM's data. Creativity in machine learning has also been explored outside of the design community. 
For example, Creative Adversarial Networks~(CAN)\cite{elgammal2017can} introduce entropy into the training to encourage the generation of surprising images. 
Readers are directed to Franceschelli~\etal's survey on the topic for more detail~\cite{franceschelli2021creativity}. 


\subsection{Evaluating Model Performance \& Benchmark Problems}
Readers may have noted that many of the works discussed have nearly identical methodologies, though they are applied to a specific dataset and only compare their work in this respect with a few other baselines. This may be attributed to a large variety of design applications compared to other domains such as computer vision and a tendency of design researchers to find solutions to their particular problem, instead of finding generalizable solutions to many.
However, this makes any measure of `state-of-the-art' algorithm dependent on specific applications. 
Without a good understanding of state-of-the-art methods that are broadly applicable across engineering domains, practitioners will struggle to select models for practical deployment. While some works in the design community have specifically introduced benchmark problems and datasets~\cite{regenwetter2021biked, regenwetter2022framed}, there is still a need for larger, higher-quality, and more numerous datasets and benchmarks. 

The difficulty of establishing a state-of-the-art lies in the lack of benchmark problems and performance metrics within the DGM field and within design automation as a whole. Design datasets tend to be small, restricted in domain, and sparse in distribution as we discuss in Section~\ref{datalimitations}. Furthermore, many methods are developed on proprietary datasets. Finally, the hugely different design representation methods across the field make establishing standardized model performance benchmarks difficult, even if there were good datasets upon which to do so. These problems are noted by many authors in the field as well, who find it difficult to compare their approach with existing ones~\cite{rawat2019application, guest_editorial}. 

\subsection{Data Limitations and Quality} \label{datalimitations}
Data sparsity is one of the greatest challenges facing the data-driven design community and is a particular concern for researchers developing data-hungry DGMs. Broadly, there are three problems when it comes to the data. The first of these is a general lack of data in many design domains. Although datasets continue to cover more and more design fields, there still are many that lack publicly available data. There are also data representations that are underrepresented in the current datasets, notably graphs as discussed in Sec.~\ref{representation}. The second key data-related problem that designers face in applying DGMs is the insufficient size of current datasets. Many of the latest breakthroughs in deep learning, notably in computer vision and natural language processing, have owed their success to very large models which are notoriously data-hungry, requiring millions or billions of training examples, and cannot be properly utilized with datasets of smaller size. Focusing resources on the generation of very large publicly available datasets or effective data augmentation methods would open doors for the design community to leverage larger deep models. The last major limitation in existing data is sparsity and bias. These problems are common in many design datasets. How the data is distributed in the performance space can cause some challenges for many DGMs, especially in inverse design problems~\cite{rangegan}. Diversity is important to ensure the data covers the design space as evenly as possible to avoid bias in the data. Keeping diversity and bias avoidance in mind will help researchers generate better datasets.

We highly encourage researchers to develop diverse datasets, publicly release these datasets, and establish benchmark problems for future works. We subsequently encourage researchers to test their methods on other researchers' data, publicly release their benchmarking results, and acknowledge state-of-the-art methods when possible.

\subsection{Other DGMs}
So far our discussion has been focused on the VAE, GAN, and RL-based approaches, which have dominated the field of DGMs in engineering design. 
Transformer-based sequential models were initially developed for natural language processing and generation or translation of text, however, researchers have shown that the scalability of these models allows for them to be used for the generation of complex data in a sequential manner~\cite{pmlr-v119-chen20s,ramesh2021zeroshot,dhariwal2020jukebox,radford2021learning}. Furthermore, these models have an unprecedented ability to be mixed with language processing. This may enable human language-based conditioning and control over the generation process of data~\cite{ramesh2021zeroshot,radford2021learning}, enabling deployment for a much wider user base and a much wider set of tasks. The applications of transformer-based models as well as their scalability may appeal to researchers in the design community. Despite this, it is important to note that these models are particularly large and therefore require very large datasets, further highlighting the need for better datasets.






\section{Conclusion}
In this review, we discussed the applications of Deep Generative Models (DGMs) across engineering design fields. To give readers a sense of the tools and methods available, we began our review with an overview of DGMs typically used for engineering design problems, emphasizing Generative Adversarial Networks (GANs), Variational Autoencoders (VAEs), and Reinforcement Learning (RL). To help weigh different design representation methods, we then discussed the strengths and weaknesses of common parameterization methods. To inform readers about existing work outside of their subdomain, we then collected and reviewed 63 papers from a variety of different engineering design subdisciplines which directly propose DGMs. To make readers aware of the datasets available on which to develop and test data-driven design methods, we review commonly-used datasets in the field. Finally, to provide inspiration, we discuss key challenges and limitations currently seen across the field and highlight possible solution pathways. 




\bibliographystyle{asmems4}

\bibliography{asme2e}

\begin{thebibliography}{100}

\bibitem{chakrabarti2011computer}
Chakrabarti, A., Shea, K., Stone, R., Cagan, J., Campbell, M., Hernandez,
  N.~V., and Wood, K.~L., 2011.
\newblock ``{Computer-Based Design Synthesis Research: An Overview}''.
\newblock {\em Journal of Computing and Information Science in Engineering,
  {\bf 11}}(2), 06.
\newblock 021003.

\bibitem{deng2014deep}
Deng, L., and Yu, D., 2014.
\newblock ``Deep learning: Methods and applications''.
\newblock {\em Found. Trends Signal Process., {\bf 7}}(3–4), jun,
  p.~197–387.

\bibitem{goodfellow2014generative}
Goodfellow, I.~J., Pouget-Abadie, J., Mirza, M., Xu, B., Warde-Farley, D.,
  Ozair, S., Courville, A., and Bengio, Y., 2014.
\newblock ``Generative adversarial nets''.
\newblock In Proceedings of the 27th International Conference on Neural
  Information Processing Systems - Volume 2, NIPS'14, MIT Press,
  p.~2672–2680.

\bibitem{CycleGAN2017}
Zhu, J.-Y., Park, T., Isola, P., and Efros, A.~A., 2017.
\newblock ``Unpaired image-to-image translation using cycle-consistent
  adversarial networks''.
\newblock In 2017 IEEE International Conference on Computer Vision (ICCV),
  pp.~2242--2251.

\bibitem{choi2020stargan}
Choi, Y., Uh, Y., Yoo, J., and Ha, J.-W., 2020.
\newblock ``Stargan v2: Diverse image synthesis for multiple domains''.
\newblock In 2020 IEEE/CVF Conference on Computer Vision and Pattern
  Recognition (CVPR), pp.~8185--8194.

\bibitem{karras2020analyzing}
Karras, T., Laine, S., Aittala, M., Hellsten, J., Lehtinen, J., and Aila, T.,
  2020.
\newblock ``Analyzing and improving the image quality of stylegan''.
\newblock In 2020 IEEE/CVF Conference on Computer Vision and Pattern
  Recognition (CVPR), pp.~8107--8116.

\bibitem{karras2019style}
Karras, T., Laine, S., and Aila, T., 2019.
\newblock ``A style-based generator architecture for generative adversarial
  networks''.
\newblock In 2019 IEEE/CVF Conference on Computer Vision and Pattern
  Recognition (CVPR), pp.~4396--4405.

\bibitem{creswell2018generative}
Creswell, A., White, T., Dumoulin, V., Arulkumaran, K., Sengupta, B., and
  Bharath, A.~A., 2018.
\newblock ``Generative adversarial networks: An overview''.
\newblock {\em IEEE Signal Processing Magazine, {\bf 35}}(1), pp.~53--65.

\bibitem{salimans2016improved}
Salimans, T., Goodfellow, I., Zaremba, W., Cheung, V., Radford, A., and Chen,
  X., 2016.
\newblock ``Improved techniques for training gans''.
\newblock In Proceedings of the 30th International Conference on Neural
  Information Processing Systems, NIPS'16, Curran Associates Inc.,
  p.~2234–2242.

\bibitem{arjovsky2017towards}
Arjovsky, M., and Bottou, L., 2017.
\newblock ``Towards principled methods for training generative adversarial
  networks''.
\newblock {\em arXiv preprint arXiv:1701.04862}.

\bibitem{arjovsky2017wasserstein}
Arjovsky, M., Chintala, S., and Bottou, L., 2017.
\newblock ``Wasserstein generative adversarial networks''.
\newblock In Proceedings of the 34th International Conference on Machine
  Learning - Volume 70, ICML'17, JMLR.org, p.~214–223.

\bibitem{gulrajani2017improved}
Gulrajani, I., Ahmed, F., Arjovsky, M., Dumoulin, V., and Courville, A., 2017.
\newblock ``Improved training of wasserstein gans''.
\newblock In Proceedings of the 31st International Conference on Neural
  Information Processing Systems, NIPS'17, Curran Associates Inc.,
  p.~5769–5779.

\bibitem{VeeGAN}
Srivastava, A., Valkov, L., Russell, C., Gutmann, M.~U., and Sutton, C., 2017.
\newblock ``Veegan: Reducing mode collapse in gans using implicit variational
  learning''.
\newblock In Advances in Neural Information Processing Systems, I.~Guyon, U.~V.
  Luxburg, S.~Bengio, H.~Wallach, R.~Fergus, S.~Vishwanathan, and R.~Garnett,
  eds., Vol.~30, Curran Associates, Inc., pp.~3308--3318.

\bibitem{improvinggan}
Salimans, T., Goodfellow, I., Zaremba, W., Cheung, V., Radford, A., Chen, X.,
  and Chen, X., 2016.
\newblock ``Improved techniques for training gans''.
\newblock In Advances in Neural Information Processing Systems, D.~Lee,
  M.~Sugiyama, U.~Luxburg, I.~Guyon, and R.~Garnett, eds., Vol.~29, Curran
  Associates, Inc., pp.~2234--2242.

\bibitem{chen2021padgan}
Chen, W., and Ahmed, F., 2021.
\newblock ``Padgan: Learning to generate high-quality novel designs''.
\newblock {\em Journal of Mechanical Design, {\bf 143}}(3), p.~031703.

\bibitem{mirza2014conditional}
Mirza, M., and Osindero, S., 2014.
\newblock ``Conditional generative adversarial nets''.
\newblock {\em arXiv preprint arXiv:1411.1784}.

\bibitem{chen2016infogan}
Chen, X., Duan, Y., Houthooft, R., Schulman, J., Sutskever, I., and Abbeel, P.,
  2016.
\newblock ``Infogan: Interpretable representation learning by information
  maximizing generative adversarial nets''.
\newblock In Proceedings of the 30th International Conference on Neural
  Information Processing Systems, pp.~2180--2188.

\bibitem{dong2019inverse}
Dong, Y., Li, D., Zhang, C., Wu, C., Wang, H., Xin, M., Cheng, J., and Lin, J.,
  2020.
\newblock ``Inverse design of two-dimensional graphene/h-bn hybrids by a
  regressional and conditional gan''.
\newblock {\em Carbon, {\bf 169}}, pp.~9--16.

\bibitem{ding2020ccgan}
Ding, X., Wang, Y., Xu, Z., Welch, W.~J., and Wang, Z.~J., 2021.
\newblock ``Ccgan: Continuous conditional generative adversarial networks for
  image generation''.
\newblock In 9th International Conference on Learning Representations, {ICLR}
  2021, Virtual Event, Austria, May 3-7, 2021, OpenReview.net.

\bibitem{pcdgan}
Heyrani~Nobari, A., Chen, W., and Ahmed, F., 2021.
\newblock ``Pcdgan: A continuous conditional diverse generative adversarial
  network for inverse design''.
\newblock {\em Proceedings of the 27th ACM SIGKDD Conference on Knowledge
  Discovery \& Data Mining}, Aug.

\bibitem{kingma2013auto}
Kingma, D.~P., and Welling, M., 2013.
\newblock ``Auto-encoding variational bayes''.
\newblock {\em arXiv preprint arXiv:1312.6114}.

\bibitem{kullback1951information}
Kullback, S., and Leibler, R.~A., 1951.
\newblock ``On information and sufficiency''.
\newblock {\em The annals of mathematical statistics, {\bf 22}}(1), pp.~79--86.

\bibitem{kingma2019introduction}
Kingma, D.~P., and Welling, M., 2019.
\newblock ``An introduction to variational autoencoders''.
\newblock {\em Foundations and Trends in Machine Learning, {\bf 12}}(4),
  pp.~307--392.

\bibitem{sohn2015learning}
Sohn, K., Yan, X., and Lee, H., 2015.
\newblock ``Learning structured output representation using deep conditional
  generative models''.
\newblock In Proceedings of the 28th International Conference on Neural
  Information Processing Systems - Volume 2, NIPS'15, MIT Press,
  p.~3483–3491.

\bibitem{kaelbling1996RL}
Kaelbling, L.~P., Littman, M.~L., and Moore, A.~W., 1996.
\newblock ``Reinforcement learning: A survey''.
\newblock {\em J. Artif. Int. Res., {\bf 4}}(1), May, p.~237–285.

\bibitem{mnih2015human}
Mnih, V., Kavukcuoglu, K., Silver, D., Rusu, A.~A., Veness, J., Bellemare,
  M.~G., Graves, A., Riedmiller, M., Fidjeland, A.~K., Ostrovski, G., et~al.,
  2015.
\newblock ``Human-level control through deep reinforcement learning''.
\newblock {\em nature, {\bf 518}}(7540), pp.~529--533.

\bibitem{daneshmand20183d}
Daneshmand, M., Helmi, A., Avots, E., Noroozi, F., Alisinanoglu, F., Arslan,
  H.~S., Gorbova, J., Haamer, R.~E., Ozcinar, C., and Anbarjafari, G., 2018.
\newblock ``3d scanning: A comprehensive survey''.
\newblock {\em arXiv preprint arXiv:1801.08863}.

\bibitem{remondino2003point}
Remondino, F., 2003.
\newblock ``From point cloud to surface: the modeling and visualization
  problem''.
\newblock {\em International Archives of the Photogrammetry, Remote Sensing and
  Spatial Information Sciences, {\bf 34}}.

\bibitem{ranjan2018generating}
Ranjan, A., Bolkart, T., Sanyal, S., and Black, M.~J., 2018.
\newblock ``Generating 3d faces using convolutional mesh autoencoders''.
\newblock In Proceedings of the European Conference on Computer Vision (ECCV),
  pp.~704--720.

\bibitem{cheng2019meshgan}
Cheng, S., Bronstein, M., Zhou, Y., Kotsia, I., Pantic, M., and Zafeiriou, S.,
  2019.
\newblock ``Meshgan: Non-linear 3d morphable models of faces''.
\newblock {\em arXiv preprint arXiv:1903.10384}.

\bibitem{zhang2020meshingnet}
Zhang, Z., Wang, Y., Jimack, P.~K., and Wang, H., 2020.
\newblock ``Meshingnet: a new mesh generation method based on deep learning''.
\newblock In International Conference on Computational Science, Springer,
  pp.~186--198.

\bibitem{wang2020deep}
Wang, L., Chan, Y.-C., Ahmed, F., Liu, Z., Zhu, P., and Chen, W., 2020.
\newblock ``Deep generative modeling for mechanistic-based learning and design
  of metamaterial systems''.
\newblock {\em Computer Methods in Applied Mechanics and Engineering, {\bf
  372}}, p.~113377.

\bibitem{chen2019synthesizing}
Chen, W., and Fuge, M., 2019.
\newblock ``Synthesizing designs with interpart dependencies using hierarchical
  generative adversarial networks''.
\newblock {\em Journal of Mechanical Design, {\bf 141}}(11), p.~111403.

\bibitem{stump2019spatial}
Stump, G.~M., Miller, S.~W., Yukish, M.~A., Simpson, T.~W., and Tucker, C.,
  2019.
\newblock ``Spatial grammar-based recurrent neural network for design form and
  behavior optimization''.
\newblock {\em Journal of Mechanical Design, {\bf 141}}(12).

\bibitem{10.1115/DETC2020-22355}
Cao, W., Robinson, T., Hua, Y., Boussuge, F., Colligan, A.~R., and Pan, W.,
  2020.
\newblock ``{Graph Representation of 3D CAD Models for Machining Feature
  Recognition With Deep Learning}''.
\newblock Vol.~Volume 11A: 46th Design Automation Conference (DAC) of {\em
  International Design Engineering Technical Conferences and Computers and
  Information in Engineering Conference}.
\newblock V11AT11A003.

\bibitem{10.1115/1.4038303}
Yang, W., Ding, H., Zi, B., and Zhang, D., 2017.
\newblock ``{New Graph Representation for Planetary Gear Trains}''.
\newblock {\em Journal of Mechanical Design, {\bf 140}}(1), 11.
\newblock 012303.

\bibitem{10.1115/1.2916916}
Hsu, C.-H., and Lam, K.-T., 1992.
\newblock ``{A New Graph Representation for the Automatic Kinematic Analysis of
  Planetary Spur-Gear Trains}''.
\newblock {\em Journal of Mechanical Design, {\bf 114}}(1), 03, pp.~196--200.

\bibitem{LEE1996831}
Lee, J.~Y., and Kim, K., 1996.
\newblock ``Geometric reasoning for knowledge-based parametric design using
  graph representation''.
\newblock {\em Computer-Aided Design, {\bf 28}}(10), pp.~831--841.

\bibitem{10.1115/DETC2014-35652}
Coatanéa, E., Nonsiri, S., Christophe, F., and Mokammel, F., 2014.
\newblock ``{Graph Based Representation and Analyses for Conceptual Stages}''.
\newblock Vol.~Volume 1A: 34th Computers and Information in Engineering
  Conference of {\em International Design Engineering Technical Conferences and
  Computers and Information in Engineering Conference}.
\newblock V01AT02A071.

\bibitem{Patalano2013AGS}
Patalano, S., Vitolo, F., and Lanzotti, A., 2013.
\newblock ``A graph-based software tool for the cad modeling of mechanical
  assemblies''.
\newblock In GRAPP/IVAPP.

\bibitem{GCN}
Henaff, M., Bruna, J., and LeCun, Y., 2015.
\newblock Deep convolutional networks on graph-structured data.

\bibitem{GTN}
Yun, S., Jeong, M., Kim, R., Kang, J., and Kim, H.~J., 2019.
\newblock ``Graph transformer networks''.
\newblock In Advances in Neural Information Processing Systems, H.~Wallach,
  H.~Larochelle, A.~Beygelzimer, F.~d\textquotesingle Alch\'{e}-Buc, E.~Fox,
  and R.~Garnett, eds., Vol.~32, Curran Associates, Inc.

\bibitem{GAT}
Veli{\v{c}}kovi{\'c}, P., Cucurull, G., Casanova, A., Romero, A., Li{\`o}, P.,
  and Bengio, Y., 2018.
\newblock ``Graph attention networks''.
\newblock In International Conference on Learning Representations.

\bibitem{vashishth2019composition}
Vashishth, S., Sanyal, S., Nitin, V., and Talukdar, P., 2019.
\newblock ``Composition-based multi-relational graph convolutional networks''.
\newblock In International Conference on Learning Representations.

\bibitem{li2017gated}
Li, Y., Tarlow, D., Brockschmidt, M., and Zemel, R., 2017.
\newblock Gated graph sequence neural networks.

\bibitem{NEURIPS2019_d0921d44}
Liao, R., Li, Y., Song, Y., Wang, S., Hamilton, W., Duvenaud, D.~K., Urtasun,
  R., and Zemel, R., 2019.
\newblock ``Efficient graph generation with graph recurrent attention
  networks''.
\newblock In Advances in Neural Information Processing Systems, H.~Wallach,
  H.~Larochelle, A.~Beygelzimer, F.~d\textquotesingle Alch\'{e}-Buc, E.~Fox,
  and R.~Garnett, eds., Vol.~32, Curran Associates, Inc.

\bibitem{bojchevski2018netgan}
Bojchevski, A., Shchur, O., Z{\"u}gner, D., and G{\"u}nnemann, S., 2018.
\newblock ``Netgan: Generating graphs via random walks''.
\newblock In International Conference on Machine Learning, pp.~609--618.

\bibitem{you2018graphrnn}
You, J., Ying, R., Ren, X., Hamilton, W., and Leskovec, J., 2018.
\newblock ``Graphrnn: Generating realistic graphs with deep auto-regressive
  models''.
\newblock In International conference on machine learning, PMLR,
  pp.~5708--5717.

\bibitem{li2018learning}
Li, Y., Vinyals, O., Dyer, C., Pascanu, R., and Battaglia, P., 2018.
\newblock Learning deep generative models of graphs.

\bibitem{decao2018molgan}
Cao, N.~D., and Kipf, T., 2018.
\newblock Molgan: An implicit generative model for small molecular graphs.

\bibitem{you2018graph}
You, J., Liu, B., Ying, R., Pande, V., and Leskovec, J., 2018.
\newblock ``Graph convolutional policy network for goal-directed molecular
  graph generation''.
\newblock In Proceedings of the 32nd International Conference on Neural
  Information Processing Systems, pp.~6412--6422.

\bibitem{sosnovik2019neural}
Sosnovik, I., and Oseledets, I., 2019.
\newblock ``Neural networks for topology optimization''.
\newblock {\em Russian Journal of Numerical Analysis and Mathematical
  Modelling, {\bf 34}}(4), pp.~215--223.

\bibitem{BEHZADI2021103014}
Behzadi, M.~M., and Ilieş, H.~T., 2021.
\newblock ``Real-time topology optimization in 3d via deep transfer learning''.
\newblock {\em Computer-Aided Design, {\bf 135}}, p.~103014.

\bibitem{KESHAVARZZADEH2021102947}
Keshavarzzadeh, V., Alirezaei, M., Tasdizen, T., and Kirby, R.~M., 2021.
\newblock ``Image-based multiresolution topology optimization using deep
  disjunctive normal shape model''.
\newblock {\em Computer-Aided Design, {\bf 130}}, p.~102947.

\bibitem{CANG201912}
Cang, R., Yao, H., and Ren, Y., 2019.
\newblock ``One-shot generation of near-optimal topology through theory-driven
  machine learning''.
\newblock {\em Computer-Aided Design, {\bf 109}}, pp.~12--21.

\bibitem{malkiel2018plasmonic}
Malkiel, I., Mrejen, M., Nagler, A., Arieli, U., Wolf, L., and Suchowski, H.,
  2018.
\newblock ``Plasmonic nanostructure design and characterization via deep
  learning''.
\newblock {\em Light: Science \& Applications, {\bf 7}}(1), pp.~1--8.

\bibitem{li2018transfer}
Li, X., Zhang, Y., Zhao, H., Burkhart, C., Brinson, L.~C., and Chen, W., 2018.
\newblock ``A transfer learning approach for microstructure reconstruction and
  structure-property predictions''.
\newblock {\em Scientific reports, {\bf 8}}(1), pp.~1--13.

\bibitem{jung2021super}
Jung, J., Na, J., Park, H.~K., Park, J.~M., Kim, G., Lee, S., and Kim, H.~S.,
  2021.
\newblock ``Super-resolving material microstructure image via deep learning for
  microstructure characterization and mechanical behavior analysis''.
\newblock {\em npj Computational Materials, {\bf 7}}(1), pp.~1--11.

\bibitem{LI2019172}
Li, B., Huang, C., Li, X., Zheng, S., and Hong, J., 2019.
\newblock ``Non-iterative structural topology optimization using deep
  learning''.
\newblock {\em Computer-Aided Design, {\bf 115}}, pp.~172--180.

\bibitem{rawat2019application}
Rawat, S., and Shen, M.~H., 2019.
\newblock Application of adversarial networks for 3d structural topology
  optimization.
\newblock Tech. rep., SAE Technical Paper.

\bibitem{oh2018design}
Oh, S., Jung, Y., Lee, I., and Kang, N., 2018.
\newblock ``Design automation by integrating generative adversarial networks
  and topology optimization''.
\newblock In International Design Engineering Technical Conferences and
  Computers and Information in Engineering Conference, Vol.~51753, American
  Society of Mechanical Engineers, p.~V02AT03A008.

\bibitem{oh2019deep}
Oh, S., Jung, Y., Kim, S., Lee, I., and Kang, N., 2019.
\newblock ``Deep generative design: Integration of topology optimization and
  generative models''.
\newblock {\em Journal of Mechanical Design, {\bf 141}}(11).

\bibitem{tan2020deep}
Tan, R.~K., Zhang, N.~L., and Ye, W., 2020.
\newblock ``A deep learning--based method for the design of microstructural
  materials''.
\newblock {\em Structural and Multidisciplinary Optimization, {\bf 61}}(4),
  pp.~1417--1438.

\bibitem{yang2018microstructural}
Yang, Z., Li, X., Catherine~Brinson, L., Choudhary, A.~N., Chen, W., and
  Agrawal, A., 2018.
\newblock ``Microstructural materials design via deep adversarial learning
  methodology''.
\newblock {\em Journal of Mechanical Design, {\bf 140}}(11).

\bibitem{ZHANG2021103041}
Zhang, H., Yang, L., Li, C., Wu, B., and Wang, W., 2021.
\newblock ``Scaffoldgan: Synthesis of scaffold materials based on generative
  adversarial networks''.
\newblock {\em Computer-Aided Design, {\bf 138}}, p.~103041.

\bibitem{mosser2017reconstruction}
Mosser, L., Dubrule, O., and Blunt, M.~J., 2017.
\newblock ``Reconstruction of three-dimensional porous media using generative
  adversarial neural networks''.
\newblock {\em Physical Review E, {\bf 96}}(4), p.~043309.

\bibitem{lee2021virtual}
Lee, J.-W., Goo, N.~H., Park, W.~B., Pyo, M., and Sohn, K.-S., 2021.
\newblock ``Virtual microstructure design for steels using generative
  adversarial networks''.
\newblock {\em Engineering Reports, {\bf 3}}(1), p.~e12274.

\bibitem{liu2019case}
Liu, S., Zhong, Z., Takbiri-Borujeni, A., Kazemi, M., Fu, Q., and Yang, Y.,
  2019.
\newblock ``A case study on homogeneous and heterogeneous reservoir porous
  media reconstruction by using generative adversarial networks''.
\newblock {\em Energy Procedia, {\bf 158}}, pp.~6164--6169.

\bibitem{shu20203d}
Shu, D., Cunningham, J., Stump, G., Miller, S.~W., Yukish, M.~A., Simpson,
  T.~W., and Tucker, C.~S., 2020.
\newblock ``3d design using generative adversarial networks and physics-based
  validation''.
\newblock {\em Journal of Mechanical Design, {\bf 142}}(7), p.~071701.

\bibitem{sharpe2019topology}
Sharpe, C., and Seepersad, C.~C., 2019.
\newblock ``Topology design with conditional generative adversarial networks''.
\newblock In International Design Engineering Technical Conferences and
  Computers and Information in Engineering Conference, Vol.~59186, American
  Society of Mechanical Engineers, p.~V02AT03A062.

\bibitem{nie2021topologygan}
Nie, Z., Lin, T., Jiang, H., and Kara, L.~B., 2021.
\newblock ``Topologygan: Topology optimization using generative adversarial
  networks based on physical fields over the initial domain''.
\newblock {\em Journal of Mechanical Design, {\bf 143}}(3), p.~031715.

\bibitem{yu2019deep}
Yu, Y., Hur, T., Jung, J., and Jang, I.~G., 2019.
\newblock ``Deep learning for determining a near-optimal topological design
  without any iteration''.
\newblock {\em Structural and Multidisciplinary Optimization, {\bf 59}}(3),
  pp.~787--799.

\bibitem{Valdez2021framework}
Valdez, S., Seepersad, C., and Kambampati, S., 2021.
\newblock ``A framework for interactive structural design exploration''.
\newblock In International Design Engineering Technical Conferences and
  Computers and Information in Engineering Conference, {IDETC-21}, ASME.

\bibitem{yilmaz2020conditional}
Yilmaz, E., and German, B., 2020.
\newblock ``Conditional generative adversarial network framework for airfoil
  inverse design''.
\newblock In AIAA aviation 2020 forum, p.~3185.

\bibitem{chen2018b}
Chen, W., and Fuge, M., 2018.
\newblock ``B{\'e}ziergan: Automatic generation of smooth curves from
  interpretable low-dimensional parameters''.
\newblock {\em arXiv preprint arXiv:1808.08871}.

\bibitem{chen2019aerodynamic}
Chen, W., Chiu, K., and Fuge, M., 2019.
\newblock ``Aerodynamic design optimization and shape exploration using
  generative adversarial networks''.
\newblock In AIAA Scitech 2019 Forum, p.~2351.

\bibitem{rangegan}
Heyrani~Nobari, A., Chen, W.~W., and Ahmed, F., 2021.
\newblock ``{RANGE-GAN: Design Synthesis Under Constraints Using Conditional
  Generative Adversarial Networks}''.
\newblock {\em Journal of Mechanical Design}, 09, pp.~1--16.

\bibitem{wang_peng_li_chen_wu_wang_childs_guo_2019}
Wang, P., Peng, D., Li, L., Chen, L., Wu, C., Wang, X., Childs, P., and Guo,
  Y., 2019.
\newblock ``Human-in-the-loop design with machine learning''.
\newblock {\em Proceedings of the Design Society: International Conference on
  Engineering Design, {\bf 1}}(1), p.~2577–2586.

\bibitem{guo2018indirect}
Guo, T., Lohan, D.~J., Cang, R., Ren, M.~Y., and Allison, J.~T., 2018.
\newblock ``An indirect design representation for topology optimization using
  variational autoencoder and style transfer''.
\newblock In 2018 AIAA/ASCE/AHS/ASC Structures, Structural Dynamics, and
  Materials Conference, p.~0804.

\bibitem{cang2018improving}
Cang, R., Li, H., Yao, H., Jiao, Y., and Ren, Y., 2018.
\newblock ``Improving direct physical properties prediction of heterogeneous
  materials from imaging data via convolutional neural network and a
  morphology-aware generative model''.
\newblock {\em Computational Materials Science, {\bf 150}}, pp.~212--221.

\bibitem{li2020designing}
Li, X., Ning, S., Liu, Z., Yan, Z., Luo, C., and Zhuang, Z., 2020.
\newblock ``Designing phononic crystal with anticipated band gap through a deep
  learning based data-driven method''.
\newblock {\em Computer Methods in Applied Mechanics and Engineering, {\bf
  361}}, p.~112737.

\bibitem{liu2020hybrid}
Liu, Z., Raju, L., Zhu, D., and Cai, W., 2020.
\newblock ``A hybrid strategy for the discovery and design of photonic
  structures''.
\newblock {\em IEEE Journal on Emerging and Selected Topics in Circuits and
  Systems, {\bf 10}}(1), pp.~126--135.

\bibitem{xue2020machine}
Xue, T., Wallin, T.~J., Menguc, Y., Adriaenssens, S., and Chiaramonte, M.,
  2020.
\newblock ``Machine learning generative models for automatic design of
  multi-material 3d printed composite solids''.
\newblock {\em Extreme Mechanics Letters, {\bf 41}}, p.~100992.

\bibitem{brock2016context}
Brock, A., Lim, T., Ritchie, J.~M., and Weston, N., 2016.
\newblock ``Context-aware content generation for virtual environments''.
\newblock In International Design Engineering Technical Conferences and
  Computers and Information in Engineering Conference, Vol.~50084, American
  Society of Mechanical Engineers, p.~V01BT02A045.

\bibitem{deshpande2019computational}
Deshpande, S., and Purwar, A., 2019.
\newblock ``Computational creativity via assisted variational synthesis of
  mechanisms using deep generative models''.
\newblock {\em Journal of Mechanical Design, {\bf 141}}(12).

\bibitem{sharma2020path}
Sharma, S., and Purwar, A., 2020.
\newblock ``Path synthesis of defect-free spatial 5-ss mechanisms using machine
  learning''.
\newblock In International Design Engineering Technical Conferences and
  Computers and Information in Engineering Conference, Vol.~83990, American
  Society of Mechanical Engineers, p.~V010T10A034.

\bibitem{regenwetter2021biked}
Regenwetter, L., Curry, B., and Ahmed, F., 2021.
\newblock ``{BIKED}: A dataset and machine learning benchmarks for data-driven
  bicycle design''.
\newblock In International Design Engineering Technical Conferences and
  Computers and Information in Engineering Conference, {IDETC-21}, ASME.

\bibitem{tang2020generative}
Tang, Y., Kojima, K., Koike-Akino, T., Wang, Y., Wu, P., Tahersima, M., Jha,
  D., Parsons, K., and Qi, M., 2020.
\newblock ``Generative deep learning model for a multi-level nano-optic
  broadband power splitter''.
\newblock In 2020 Optical Fiber Communications Conference and Exhibition (OFC),
  IEEE, pp.~1--3.

\bibitem{chen2021geometry}
Chen, H., and Liu, X., 2021.
\newblock ``Geometry enhanced generative adversarial networks for random
  heterogeneous material representation''.
\newblock In International Design Engineering Technical Conferences and
  Computers and Information in Engineering Conference, {IDETC-21}, ASME.

\bibitem{burnap2016estimating}
Burnap, A., Liu, Y., Pan, Y., Lee, H., Gonzalez, R., and Papalambros, P.~Y.,
  2016.
\newblock ``Estimating and exploring the product form design space using deep
  generative models''.
\newblock In International Design Engineering Technical Conferences and
  Computers and Information in Engineering Conference, Vol.~50107, American
  Society of Mechanical Engineers, p.~V02AT03A013.

\bibitem{ma2019probabilistic}
Ma, W., Cheng, F., Xu, Y., Wen, Q., and Liu, Y., 2019.
\newblock ``Probabilistic representation and inverse design of metamaterials
  based on a deep generative model with semi-supervised learning strategy''.
\newblock {\em Advanced Materials, {\bf 31}}(35), p.~1901111.

\bibitem{zhang20193d}
Zhang, W., Yang, Z., Jiang, H., Nigam, S., Yamakawa, S., Furuhata, T., Shimada,
  K., and Kara, L.~B., 2019.
\newblock ``3d shape synthesis for conceptual design and optimization using
  variational autoencoders''.
\newblock In International Design Engineering Technical Conferences and
  Computers and Information in Engineering Conference, Vol.~59186, American
  Society of Mechanical Engineers, p.~V02AT03A017.

\bibitem{10.1115/1.4048422}
Deshpande, S., and Purwar, A., 2020.
\newblock ``{An Image-Based Approach to Variational Path Synthesis of
  Linkages}''.
\newblock {\em Journal of Computing and Information Science in Engineering,
  {\bf 21}}(2), 10.
\newblock 021005.

\bibitem{li2021learning}
Li, R., Zhang, Y., and Chen, H., 2021.
\newblock ``Learning the aerodynamic design of supercritical airfoils through
  deep reinforcement learning''.
\newblock {\em AIAA Journal}, pp.~1--14.

\bibitem{dering2018physics}
Dering, M., Cunningham, J., Desai, R., Yukish, M.~A., Simpson, T.~W., and
  Tucker, C.~S., 2018.
\newblock ``A physics-based virtual environment for enhancing the quality of
  deep generative designs''.
\newblock In International Design Engineering Technical Conferences and
  Computers and Information in Engineering Conference, Vol.~51753, American
  Society of Mechanical Engineers, p.~V02AT03A015.

\bibitem{lee2019case}
Lee, X.~Y., Balu, A., Stoecklein, D., Ganapathysubramanian, B., and Sarkar, S.,
  2019.
\newblock ``A case study of deep reinforcement learning for engineering design:
  Application to microfluidic devices for flow sculpting''.
\newblock {\em Journal of Mechanical Design, {\bf 141}}(11), p.~111401.

\bibitem{raina2021goal}
Raina, A., Puentes, L., Cagan, J., and McComb, C., 2021.
\newblock ``Goal-directed design agents: Integrating visual imitation with
  one-step lookahead optimization for generative design''.
\newblock {\em Journal of Mechanical Design, {\bf 143}}(12), p.~124501.

\bibitem{10.1115/DETC2019-97711}
Lopez, C.~E., Ashour, O., and Tucker, C.~S., 2019.
\newblock ``{Reinforcement Learning Content Generation for Virtual Reality
  Applications}''.
\newblock Vol.~Volume 1: 39th Computers and Information in Engineering
  Conference of {\em International Design Engineering Technical Conferences and
  Computers and Information in Engineering Conference}.
\newblock V001T02A009.

\bibitem{10.1115/DETC2020-22624}
Cunningham, J., Lopez, C., Ashour, O., and Tucker, C.~S., 2020.
\newblock ``{Multi-Context Generation in Virtual Reality Environments Using
  Deep Reinforcement Learning}''.
\newblock Vol.~Volume 9: 40th Computers and Information in Engineering
  Conference (CIE) of {\em International Design Engineering Technical
  Conferences and Computers and Information in Engineering Conference}.
\newblock V009T09A072.

\bibitem{greminger2020generative}
Greminger, M., 2020.
\newblock ``Generative adversarial networks with synthetic training data for
  enforcing manufacturing constraints on topology optimization''.
\newblock In International Design Engineering Technical Conferences and
  Computers and Information in Engineering Conference, Vol.~84003, American
  Society of Mechanical Engineers, p.~V11AT11A005.

\bibitem{Fujita2021Design}
Fujita, K., Minowa, K., Nomaguchi, Y., Yamasaki, S., and Yaji, K., 2021.
\newblock ``Design concept generation with variational deep embedding over
  comprehensive optimization''.
\newblock In International Design Engineering Technical Conferences and
  Computers and Information in Engineering Conference, {IDETC-21}, ASME.

\bibitem{cang2017scalable}
Cang, R., Vipradas, A., and Ren, Y., 2017.
\newblock ``Scalable microstructure reconstruction with multi-scale pattern
  preservation''.
\newblock In International Design Engineering Technical Conferences and
  Computers and Information in Engineering Conference, Vol.~58134, American
  Society of Mechanical Engineers, p.~V02BT03A010.

\bibitem{cang2017microstructure}
Cang, R., Xu, Y., Chen, S., Liu, Y., Jiao, Y., and Yi~Ren, M., 2017.
\newblock ``Microstructure representation and reconstruction of heterogeneous
  materials via deep belief network for computational material design''.
\newblock {\em Journal of Mechanical Design, {\bf 139}}(7), p.~071404.

\bibitem{fokina2020microstructure}
Fokina, D., Muravleva, E., Ovchinnikov, G., and Oseledets, I., 2020.
\newblock ``Microstructure synthesis using style-based generative adversarial
  networks''.
\newblock {\em Physical Review E, {\bf 101}}(4), p.~043308.

\bibitem{Wang2021gaussian}
Wang, Z., Xian, W., Baccouche, M.~R., Lanzerath, H., Li, Y., and Xu, H., 2021.
\newblock ``A gaussian mixture variational autoencoder-based approach for
  designing phononic bandgap metamaterials''.
\newblock In International Design Engineering Technical Conferences and
  Computers and Information in Engineering Conference, {IDETC-21}, ASME.

\bibitem{10.1115/DETC2016-59404}
Cang, R., and Ren, M.~Y., 2016.
\newblock ``{Deep Network-Based Feature Extraction and Reconstruction of
  Complex Material Microstructures}''.
\newblock Vol.~Volume 2B: 42nd Design Automation Conference of {\em
  International Design Engineering Technical Conferences and Computers and
  Information in Engineering Conference}.
\newblock V02BT03A008.

\bibitem{10.1115/DETC2018-85529}
Vermeer, K., Kuppens, R., and Herder, J., 2018.
\newblock ``{Kinematic Synthesis Using Reinforcement Learning}''.
\newblock Vol.~Volume 2A: 44th Design Automation Conference of {\em
  International Design Engineering Technical Conferences and Computers and
  Information in Engineering Conference}.
\newblock V02AT03A009.

\bibitem{raina2019learning}
Raina, A., McComb, C., and Cagan, J., 2019.
\newblock ``Learning to design from humans: Imitating human designers through
  deep learning''.
\newblock {\em Journal of Mechanical Design, {\bf 141}}(11).

\bibitem{puentes2020modeling}
Puentes, L., Raina, A., Cagan, J., and McComb, C., 2020.
\newblock ``Modeling a strategic human engineering design process:
  Human-inspired heuristic guidance through learned visual design agents''.
\newblock In Proceedings of the Design Society: DESIGN Conference, Vol.~1,
  Cambridge University Press, pp.~355--364.

\bibitem{yukish2020using}
Yukish, M.~A., Stump, G.~M., and Miller, S.~W., 2020.
\newblock ``Using recurrent neural networks to model spatial grammars for
  design creation''.
\newblock {\em Journal of Mechanical Design, {\bf 142}}(10), p.~104501.

\bibitem{zhu2016topology}
Zhu, J.-H., Zhang, W.-H., and Xia, L., 2016.
\newblock ``Topology optimization in aircraft and aerospace structures
  design''.
\newblock {\em Archives of Computational Methods in Engineering, {\bf 23}}(4),
  pp.~595--622.

\bibitem{xia2017recent}
Xia, L., and Breitkopf, P., 2017.
\newblock ``Recent advances on topology optimization of multiscale nonlinear
  structures''.
\newblock {\em Archives of Computational Methods in Engineering, {\bf 24}}(2),
  pp.~227--249.

\bibitem{borrvall2003topology}
Borrvall, T., and Petersson, J., 2003.
\newblock ``Topology optimization of fluids in stokes flow''.
\newblock {\em International journal for numerical methods in fluids, {\bf
  41}}(1), pp.~77--107.

\bibitem{zhou2008variational}
Zhou, S., and Li, Q., 2008.
\newblock ``A variational level set method for the topology optimization of
  steady-state navier--stokes flow''.
\newblock {\em Journal of Computational Physics, {\bf 227}}(24),
  pp.~10178--10195.

\bibitem{zegard2016bridging}
Zegard, T., and Paulino, G.~H., 2016.
\newblock ``Bridging topology optimization and additive manufacturing''.
\newblock {\em Structural and Multidisciplinary Optimization, {\bf 53}}(1),
  pp.~175--192.

\bibitem{langelaar2016topology}
Langelaar, M., 2016.
\newblock ``Topology optimization of 3d self-supporting structures for additive
  manufacturing''.
\newblock {\em Additive Manufacturing, {\bf 12}}, pp.~60--70.

\bibitem{dbouk2017review}
Dbouk, T., 2017.
\newblock ``A review about the engineering design of optimal heat transfer
  systems using topology optimization''.
\newblock {\em Applied Thermal Engineering, {\bf 112}}, pp.~841--854.

\bibitem{koga2013development}
Koga, A.~A., Lopes, E. C.~C., Nova, H. F.~V., De~Lima, C.~R., and Silva, E.
  C.~N., 2013.
\newblock ``Development of heat sink device by using topology optimization''.
\newblock {\em International Journal of Heat and Mass Transfer, {\bf 64}},
  pp.~759--772.

\bibitem{gatys2015texture}
Gatys, L., Ecker, A.~S., and Bethge, M., 2015.
\newblock ``Texture synthesis using convolutional neural networks''.
\newblock {\em Advances in neural information processing systems, {\bf 28}},
  pp.~262--270.

\bibitem{gatys2016image}
Gatys, L.~A., Ecker, A.~S., and Bethge, M., 2016.
\newblock ``Image style transfer using convolutional neural networks''.
\newblock In Proceedings of the IEEE conference on computer vision and pattern
  recognition, pp.~2414--2423.

\bibitem{berthelot2017began}
Berthelot, D., Schumm, T., and Metz, L., 2017.
\newblock ``Began: Boundary equilibrium generative adversarial networks''.
\newblock {\em arXiv preprint arXiv:1703.10717}.

\bibitem{jiang2017variational}
Jiang, Z., Zheng, Y., Tan, H., Tang, B., and Zhou, H., 2017.
\newblock ``Variational deep embedding: an unsupervised and generative approach
  to clustering''.
\newblock In Proceedings of the 26th International Joint Conference on
  Artificial Intelligence, pp.~1965--1972.

\bibitem{ledig2017photo}
Ledig, C., Theis, L., Husz{\'a}r, F., Caballero, J., Cunningham, A., Acosta,
  A., Aitken, A., Tejani, A., Totz, J., Wang, Z., et~al., 2017.
\newblock ``Photo-realistic single image super-resolution using a generative
  adversarial network''.
\newblock In 2017 IEEE Conference on Computer Vision and Pattern Recognition
  (CVPR), IEEE, pp.~105--114.

\bibitem{isola2017image}
Isola, P., Zhu, J.-Y., Zhou, T., and Efros, A.~A., 2017.
\newblock ``Image-to-image translation with conditional adversarial networks''.
\newblock In Proceedings of the IEEE conference on computer vision and pattern
  recognition, pp.~1125--1134.

\bibitem{he2016deep}
He, K., Zhang, X., Ren, S., and Sun, J., 2016.
\newblock ``Deep residual learning for image recognition''.
\newblock In Proceedings of the IEEE conference on computer vision and pattern
  recognition, pp.~770--778.

\bibitem{hu2018squeeze}
Hu, J., Shen, L., and Sun, G., 2018.
\newblock ``Squeeze-and-excitation networks''.
\newblock In Proceedings of the IEEE conference on computer vision and pattern
  recognition, pp.~7132--7141.

\bibitem{long2015fully}
Long, J., Shelhamer, E., and Darrell, T., 2015.
\newblock ``Fully convolutional networks for semantic segmentation''.
\newblock In Proceedings of the IEEE conference on computer vision and pattern
  recognition, pp.~3431--3440.

\bibitem{bostanabad2018computational}
Bostanabad, R., Zhang, Y., Li, X., Kearney, T., Brinson, L.~C., Apley, D.~W.,
  Liu, W.~K., and Chen, W., 2018.
\newblock ``Computational microstructure characterization and reconstruction:
  Review of the state-of-the-art techniques''.
\newblock {\em Progress in Materials Science, {\bf 95}}, pp.~1--41.

\bibitem{lee2009convolutional}
Lee, H., Grosse, R., Ranganath, R., and Ng, A.~Y., 2009.
\newblock ``Convolutional deep belief networks for scalable unsupervised
  learning of hierarchical representations''.
\newblock In Proceedings of the 26th annual international conference on machine
  learning, pp.~609--616.

\bibitem{yu2017characterization}
Yu, S., Zhang, Y., Wang, C., Lee, W.-k., Dong, B., Odom, T.~W., Sun, C., and
  Chen, W., 2017.
\newblock ``Characterization and design of functional quasi-random
  nanostructured materials using spectral density function''.
\newblock {\em Journal of Mechanical Design, {\bf 139}}(7), p.~071401.

\bibitem{molesky2018inverse}
Molesky, S., Lin, Z., Piggott, A.~Y., Jin, W., Vuckovi{\'c}, J., and Rodriguez,
  A.~W., 2018.
\newblock ``Inverse design in nanophotonics''.
\newblock {\em Nature Photonics, {\bf 12}}(11), pp.~659--670.

\bibitem{dilokthanakul2016deep}
Dilokthanakul, N., Mediano, P.~A., Garnelo, M., Lee, M.~C., Salimbeni, H.,
  Arulkumaran, K., and Shanahan, M., 2016.
\newblock ``Deep unsupervised clustering with gaussian mixture variational
  autoencoders''.
\newblock {\em arXiv preprint arXiv:1611.02648}.

\bibitem{schulman2017proximal}
Schulman, J., Wolski, F., Dhariwal, P., Radford, A., and Klimov, O., 2017.
\newblock ``Proximal policy optimization algorithms''.
\newblock {\em arXiv preprint arXiv:1707.06347}.

\bibitem{gielis2003generic}
Gielis, J., 2003.
\newblock ``A generic geometric transformation that unifies a wide range of
  natural and abstract shapes''.
\newblock {\em American journal of botany, {\bf 90}}(3), pp.~333--338.

\bibitem{ha2018neural}
Ha, D., and Eck, D., 2018.
\newblock ``A neural representation of sketch drawings''.
\newblock In International Conference on Learning Representations.

\bibitem{kulesza2012determinantal}
Kulesza, A., Taskar, B., et~al., 2012.
\newblock ``Determinantal point processes for machine learning''.
\newblock {\em Foundations and Trends{\textregistered} in Machine Learning,
  {\bf 5}}(2--3), pp.~123--286.

\bibitem{chen2021mopadgan}
Chen, W., and Ahmed, F., 2021.
\newblock ``Mo-padgan: Reparameterizing engineering designs for augmented
  multi-objective optimization''.
\newblock {\em Applied Soft Computing, {\bf 113}}, p.~107909.

\bibitem{li2017grass}
Li, J., Xu, K., Chaudhuri, S., Yumer, E., Zhang, H., and Guibas, L., 2017.
\newblock ``Grass: Generative recursive autoencoders for shape structures''.
\newblock {\em ACM Transactions on Graphics (TOG), {\bf 36}}(4), pp.~1--14.

\bibitem{zou20173d}
Zou, C., Yumer, E., Yang, J., Ceylan, D., and Hoiem, D., 2017.
\newblock ``3d-prnn: Generating shape primitives with recurrent neural
  networks''.
\newblock In Proceedings of the IEEE International Conference on Computer
  Vision, pp.~900--909.

\bibitem{groueix2018papier}
Groueix, T., Fisher, M., Kim, V.~G., Russell, B.~C., and Aubry, M., 2018.
\newblock ``A papier-m{\^a}ch{\'e} approach to learning 3d surface
  generation''.
\newblock In Proceedings of the IEEE conference on computer vision and pattern
  recognition, pp.~216--224.

\bibitem{gao2019sdm}
Gao, L., Yang, J., Wu, T., Yuan, Y.-J., Fu, H., Lai, Y.-K., and Zhang, H.,
  2019.
\newblock ``Sdm-net: Deep generative network for structured deformable mesh''.
\newblock {\em ACM Transactions on Graphics (TOG), {\bf 38}}(6), pp.~1--15.

\bibitem{mo2019structurenet}
Mo, K., Guerrero, P., Yi, L., Su, H., Wonka, P., Mitra, N.~J., and Guibas,
  L.~J., 2019.
\newblock ``Structurenet: hierarchical graph networks for 3d shape
  generation''.
\newblock {\em ACM Transactions on Graphics, {\bf 38}}(6), pp.~1--19.

\bibitem{park2019deepsdf}
Park, J.~J., Florence, P., Straub, J., Newcombe, R., and Lovegrove, S., 2019.
\newblock ``Deepsdf: Learning continuous signed distance functions for shape
  representation''.
\newblock In Proceedings of the IEEE/CVF Conference on Computer Vision and
  Pattern Recognition, pp.~165--174.

\bibitem{liu2018learning}
Liu, S., Giles, L., and Ororbia, A., 2018.
\newblock ``Learning a hierarchical latent-variable model of 3d shapes''.
\newblock In 2018 International Conference on 3D Vision (3DV), IEEE,
  pp.~542--551.

\bibitem{qi2017pointnet}
Qi, C.~R., Su, H., Mo, K., and Guibas, L.~J., 2017.
\newblock ``Pointnet: Deep learning on point sets for 3d classification and
  segmentation''.
\newblock In Proceedings of the IEEE conference on computer vision and pattern
  recognition, pp.~652--660.

\bibitem{karnewar2020msg}
Karnewar, A., and Wang, O., 2020.
\newblock ``Msg-gan: Multi-scale gradients for generative adversarial
  networks''.
\newblock In Proceedings of the IEEE/CVF Conference on Computer Vision and
  Pattern Recognition, pp.~7799--7808.

\bibitem{van2016deep}
van Hasselt, H., Guez, A., and Silver, D., 2016.
\newblock ``Deep reinforcement learning with double q-learning''.
\newblock {\em Proceedings of the AAAI Conference on Artificial Intelligence,
  {\bf 30}}(1), Mar.

\bibitem{odena2017conditional}
Odena, A., Olah, C., and Shlens, J., 2017.
\newblock ``Conditional image synthesis with auxiliary classifier gans''.
\newblock In International conference on machine learning, PMLR,
  pp.~2642--2651.

\bibitem{mccomb2018data}
McComb, C., Cagan, J., and Kotovsky, K., 2018.
\newblock ``Data on the design of truss structures by teams of engineering
  students''.
\newblock {\em Data in brief, {\bf 18}}, pp.~160--163.

\bibitem{regenwetter2022biked}
Regenwetter, L., Curry, B., and Ahmed, F., 2022.
\newblock ``Biked: A dataset for computational bicycle design with machine
  learning benchmarks''.
\newblock {\em Journal of Mechanical Design, {\bf 144}}(3).

\bibitem{hunter2017topy}
Hunter, W., et~al., 2017.
\newblock Topy-topology optimization with python.

\bibitem{iren2021aachen}
Iren, D., Ackermann, M., Gorfer, J., Pujar, G., Wesselmecking, S., Krupp, U.,
  and Bromuri, S., 2021.
\newblock ``Aachen-heerlen annotated steel microstructure dataset''.
\newblock {\em Scientific Data, {\bf 8}}(1), pp.~1--9.

\bibitem{larmuseau2020compact}
Larmuseau, M., Sluydts, M., Theuwissen, K., Duprez, L., Dhaene, T., and
  Cottenier, S., 2020.
\newblock ``Compact representations of microstructure images using triplet
  networks''.
\newblock {\em npj Computational Materials, {\bf 6}}(1), pp.~1--11.

\bibitem{decost2017uhcsdb}
DeCost, B.~L., Hecht, M.~D., Francis, T., Webler, B.~A., Picard, Y.~N., and
  Holm, E.~A., 2017.
\newblock ``Uhcsdb: ultrahigh carbon steel micrograph database''.
\newblock {\em Integrating Materials and Manufacturing Innovation, {\bf 6}}(2),
  pp.~197--205.

\bibitem{zhao2018nanomine}
Zhao, H., Wang, Y., Lin, A., Hu, B., Yan, R., McCusker, J., Chen, W.,
  McGuinness, D.~L., Schadler, L., and Brinson, L.~C., 2018.
\newblock ``Nanomine schema: An extensible data representation for polymer
  nanocomposites''.
\newblock {\em APL Materials, {\bf 6}}(11), p.~111108.

\bibitem{chang2015shapenet}
Chang, A.~X., Funkhouser, T., Guibas, L., Hanrahan, P., Huang, Q., Li, Z.,
  Savarese, S., Savva, M., Song, S., Su, H., et~al., 2015.
\newblock ``Shapenet: An information-rich 3d model repository''.
\newblock {\em arXiv preprint arXiv:1512.03012}.

\bibitem{mo2019partnet}
Mo, K., Zhu, S., Chang, A.~X., Yi, L., Tripathi, S., Guibas, L.~J., and Su, H.,
  2019.
\newblock ``Partnet: A large-scale benchmark for fine-grained and hierarchical
  part-level 3d object understanding''.
\newblock In Proceedings of the IEEE/CVF Conference on Computer Vision and
  Pattern Recognition, pp.~909--918.

\bibitem{wu20153d}
Wu, Z., Song, S., Khosla, A., Yu, F., Zhang, L., Tang, X., and Xiao, J., 2015.
\newblock ``3d shapenets: A deep representation for volumetric shapes''.
\newblock In Proceedings of the IEEE conference on computer vision and pattern
  recognition, pp.~1912--1920.

\bibitem{sangpil2020large}
Kim, S., Chi, H.-g., Hu, X., Huang, Q., and Ramani, K., 2020.
\newblock ``A large-scale annotated mechanical components benchmark for
  classification and retrieval tasks with deep neural networks''.
\newblock In Proceedings of 16th European Conference on Computer Vision (ECCV).

\bibitem{willis2021fusion}
Willis, K.~D., Pu, Y., Luo, J., Chu, H., Du, T., Lambourne, J.~G.,
  Solar-Lezama, A., and Matusik, W., 2021.
\newblock ``Fusion 360 gallery: A dataset and environment for programmatic cad
  construction from human design sequences''.
\newblock {\em ACM Transactions on Graphics (TOG), {\bf 40}}(4), pp.~1--24.

\bibitem{nobari2021creativegan}
Nobari, A.~H., Rashad, M.~F., and Ahmed, F., 2021.
\newblock ``Creativegan: Editing generative adversarial networks for creative
  design synthesis''.
\newblock In International Design Engineering Technical Conferences and
  Computers and Information in Engineering Conference, {IDETC-21}, ASME.

\bibitem{regenwetter2022framed}
Regenwetter, L., Weaver, C., and Ahmed, F., 2022.
\newblock Framed: Data-driven structural performance analysis of
  community-designed bicycle frames.

\bibitem{chan2021metaset}
Chan, Y.-C., Ahmed, F., Wang, L., and Chen, W., 2021.
\newblock ``Metaset: Exploring shape and property spaces for data-driven
  metamaterials design''.
\newblock {\em Journal of Mechanical Design, {\bf 143}}(3), p.~031707.

\bibitem{wang2021data}
Wang, L., van Beek, A., Da, D., Chan, Y.-C., Zhu, P., and Chen, W., 2021.
\newblock ``Data-driven multiscale design of cellular composites with
  multiclass microstructures for natural frequency maximization''.
\newblock {\em arXiv preprint arXiv:2106.06478}.

\bibitem{jongejan2016quick}
Jongejan, J., Rowley, H., Kawashima, T., Kim, J., and Fox-Gieg, N., 2016.
\newblock ``The quick, draw!-ai experiment''.
\newblock {\em Mount View, CA, accessed Feb, {\bf 17}}(2018), p.~4.

\bibitem{lopez2018human}
Lopez, C., Miller, S.~R., and Tucker, C.~S., 2018.
\newblock ``Human validation of computer vs human generated design sketches''.
\newblock In International Design Engineering Technical Conferences and
  Computers and Information in Engineering Conference, Vol.~51845, American
  Society of Mechanical Engineers, p.~V007T06A015.

\bibitem{toh2013exploring}
Toh, C.~A., and Miller, S.~R., 2013.
\newblock ``Exploring the utility of product dissection for early-phase idea
  generation''.
\newblock In International Design Engineering Technical Conferences and
  Computers and Information in Engineering Conference, Vol.~55928, American
  Society of Mechanical Engineers, p.~V005T06A034.

\bibitem{FEA1}
Liang, L., Liu, M., Martin, C., and Sun, W., 2018.
\newblock ``A deep learning approach to estimate stress distribution: a fast
  and accurate surrogate of finite-element analysis''.
\newblock {\em Journal of The Royal Society Interface, {\bf 15}}(138),
  p.~20170844.

\bibitem{FEA2}
Jiang, H., Nie, Z., Yeo, R., Farimani, A.~B., and Kara, L.~B., 2020.
\newblock ``{StressGAN: A Generative Deep Learning Model for 2D Stress
  Distribution Prediction}''.
\newblock Vol.~Volume 11B: 46th Design Automation Conference (DAC) of {\em
  International Design Engineering Technical Conferences and Computers and
  Information in Engineering Conference}.
\newblock V11BT11A023.

\bibitem{FEA3}
Nie, Z., Jiang, H., and Kara, L.~B., 2019.
\newblock ``{Stress Field Prediction in Cantilevered Structures Using
  Convolutional Neural Networks}''.
\newblock Vol.~Volume 1: 39th Computers and Information in Engineering
  Conference of {\em International Design Engineering Technical Conferences and
  Computers and Information in Engineering Conference}.
\newblock V001T02A011.

\bibitem{FEA8}
Nie, Z., Jiang, H., and Kara, L.~B., 2019.
\newblock ``Stress field prediction in cantilevered structures using
  convolutional neural networks''.
\newblock {\em Journal of Computing and Information Science in Engineering,
  {\bf 20}}(1), Sep.

\bibitem{CFD10}
Pfaff, T., Fortunato, M., Sanchez-Gonzalez, A., and Battaglia, P., 2020.
\newblock ``Learning mesh-based simulation with graph networks''.
\newblock In International Conference on Learning Representations.

\bibitem{CFD1}
Kochkov, D., Smith, J.~A., Alieva, A., Wang, Q., Brenner, M.~P., and Hoyer, S.,
  2021.
\newblock ``Machine learning–accelerated computational fluid dynamics''.
\newblock {\em Proceedings of the National Academy of Sciences, {\bf 118}}(21).

\bibitem{CFD3}
Duraisamy, K., Iaccarino, G., and Xiao, H., 2019.
\newblock ``Turbulence modeling in the age of data''.
\newblock {\em Annual Review of Fluid Mechanics, {\bf 51}}(1), pp.~357--377.

\bibitem{CFD5}
Kim, B., Azevedo, V.~C., Thuerey, N., Kim, T., Gross, M., and Solenthaler, B.,
  2019.
\newblock ``Deep fluids: A generative network for parameterized fluid
  simulations''.
\newblock {\em Computer Graphics Forum, {\bf 38}}(2), May, p.~59–70.

\bibitem{dering2017generative}
Dering, M.~L., and Tucker, C.~S., 2017.
\newblock ``Generative adversarial networks for increasing the veracity of big
  data''.
\newblock In 2017 IEEE International Conference on Big Data (Big Data), IEEE,
  pp.~2595--2602.

\bibitem{guest_editorial}
Panchal, J.~H., Fuge, M., Liu, Y., Missoum, S., and Tucker, C., 2019.
\newblock ``{Special Issue: Machine Learning for Engineering Design}''.
\newblock {\em Journal of Mechanical Design, {\bf 141}}(11), 10.
\newblock 110301.

\bibitem{elgammal2017can}
Elgammal, A., Liu, B., Elhoseiny, M., and Mazzone, M., 2017.
\newblock ``Can: Creative adversarial networks generating “art” by learning
  about styles and deviating from style norms''.
\newblock In 8th International Conference on Computational Creativity, ICCC
  2017, Georgia Institute of Technology.

\bibitem{franceschelli2021creativity}
Franceschelli, G., and Musolesi, M., 2021.
\newblock Creativity and machine learning: A survey.

\bibitem{pmlr-v119-chen20s}
Chen, M., Radford, A., Child, R., Wu, J., Jun, H., Luan, D., and Sutskever, I.,
  2020.
\newblock ``Generative pretraining from pixels''.
\newblock In Proceedings of the 37th International Conference on Machine
  Learning, H.~D. III and A.~Singh, eds., Vol.~119 of {\em Proceedings of
  Machine Learning Research}, PMLR, pp.~1691--1703.

\bibitem{ramesh2021zeroshot}
Ramesh, A., Pavlov, M., Goh, G., Gray, S., Voss, C., Radford, A., Chen, M., and
  Sutskever, I., 2021.
\newblock Zero-shot text-to-image generation.

\bibitem{dhariwal2020jukebox}
Dhariwal, P., Jun, H., Payne, C., Kim, J.~W., Radford, A., and Sutskever, I.,
  2020.
\newblock Jukebox: A generative model for music.

\bibitem{radford2021learning}
Radford, A., Kim, J.~W., Hallacy, C., Ramesh, A., Goh, G., Agarwal, S., Sastry,
  G., Askell, A., Mishkin, P., Clark, J., Krueger, G., and Sutskever, I., 2021.
\newblock Learning transferable visual models from natural language
  supervision.

\end{thebibliography}

\appendix       

\end{document}